\documentclass[letterpaper]{article} 
\usepackage{aaai23}  
\usepackage{times}  
\usepackage{helvet}  
\usepackage{courier}  
\usepackage[hyphens]{url}  
\usepackage{graphicx} 
\usepackage{natbib}  
\usepackage{caption} 
\usepackage{authblk}

\usepackage{algorithm}
\usepackage{algorithmic}

\usepackage{newfloat}
\usepackage{listings}
\DeclareCaptionStyle{ruled}{labelfont=normalfont,labelsep=colon,strut=off} 
\floatstyle{ruled}
\newfloat{listing}{tb}{lst}{}
\floatname{listing}{Listing}
\usepackage{bibentry}
\usepackage[math]{alp}
\usepackage{wrapfig}
\usepackage{subcaption}
\usepackage{rotating}
\usepackage{multirow}
\usepackage{booktabs}       
\usepackage{amsfonts}       
\usepackage{nicefrac}       
\usepackage{microtype}      
\usepackage{xcolor}         
\usepackage{graphicx}
\usepackage{amsthm}

\newtheorem{theorem}{Theorem}

\newtheorem*{proof*}{Proof}

\title{NVDiff: Graph Generation through the Diffusion of Node Vectors}
\author{ Xiaohui Chen$^1$,
    Yukun Li$^1$,
    Aonan Zhang$^2$,
    Li-Ping Liu$^{1}$\footnote{Corresponding email: liping.liu@tufts.edu}\\
}

\affiliations{
    $^1$Department of Computer Science, Tufts University\\
    $^2$ByteDance Inc.
}
\begin{document}

\maketitle


\begin{abstract}


Learning to generate graphs is challenging as a graph is a set of pairwise connected, unordered nodes encoding complex combinatorial structures. Recently, several works have proposed graph generative models based on normalizing flows or score-based diffusion models. However, these models need to generate nodes and edges in parallel from the same process, whose dimensionality is unnecessarily high. We propose NVDiff, which takes the VGAE structure and uses a score-based generative model (SGM) as a flexible prior to sample node vectors. By modeling only node vectors in the latent space, NVDiff significantly reduces the dimension of the diffusion process and thus improves sampling speed. Built on the NVDiff framework, we introduce an attention-based score network capable of capturing both local and global contexts of graphs. Experiments indicate that NVDiff significantly reduces computations and can model much larger graphs than competing methods. At the same time, it achieves superior or competitive performances over various datasets compared to previous methods. 

\end{abstract}

\section{Introduction}\label{sec:intro}

Graph generation learns a distribution over graphs that helps summarize, predict, and explore various domains such as molecules and social networks~\citep{jin2018junction,watts1998collective}. This task is challenging because it requires a generative model to learn complex combinatorial patterns hidden in the graph data. One category of neural generative models sequentially generates nodes and edges of a graph~\citep{you2018graphrnn,liao2019efficient}. However, the generation probability of such a model varies when node ordering changes. Fitting such a model often requires node orderings, which often do not exist in training data. Furthermore, it is hard to train such models by maximizing the data likelihood~\citep{chen2021order}.

This paper focuses on another category of graph generative models, which assumes nodes are exchangeable~\citep{aldous1981representations,hoover1979}. Those models generate graphs as adjacency matrices invariant in probability when jointly shuffling rows and columns with a permutation $\pi$. There are two benefits to using exchangeable graphs. First, since node ordering does not affect the generation probability, calculating the likelihood of the graph only requires one pass with an arbitrary node ordering. Second, graph generation can be done in one-shot, i.e., parallel rather than sequential.


One early example of deep exchangeable graph generative models is VGAE ~\citep{kipf2016variational}. Based on the variational auto-encoder (VAE) framework~\cite{kingma2013auto}, VGAE represents the graph by latent node vectors. The priors of those latent vectors follow a multivariate Gaussian distribution, and they use an encoder to infer the variational posterior. VGAE uses an over-simplified prior and decoder. Its node vectors are independent, thus they cannot ``coordinate'' to play different roles in a meaningful graph.  


Recently, normalizing flows (NF)~\citep{rezende2015variational,liu2019graph,lippe2020categorical,xu2021learning} are introduced for graph generation by learning a reversible mapping from a Gaussian distribution to a continuous latent graph representation. However, the reversibility assumption restricts model design for NF models. For instance, graph normalizing flows (GNF)~\citep{liu2019graph} detaches NF with Graph Auto-Encoders, thus not guaranteed to maximize the overall likelihood. GraphCNF~\citep{lippe2020categorical} incorporates both node and edge latent vectors in NF and thus has a generative space with large dimensionality. 



Score-based generative models (SGMs) introduce a diffusion process that maps observations into a noise distribution—reversing the diffusion process results in a generative model, which can be learned by score matching using, e.g., noise conditional score networks (NCSN)~\citep{song2019generative}. SGMs demonstrate superior performance than NF in generating realistic images and are recently adapted for one-shot graph generation. \citet{niu2020permutation} proposed EDP-GNN to model graphs using discrete-time SGMs. \citet{jo2022score} introduced GDSS, a continuous-time graph diffusion process for nodes and edges. Learning the graph generative process requires solving a system of reverse-time SDEs~\citep{anderson1982reverse,song2020score}. Both EDP-GNN and GDSS directly apply diffusion processes on edges. These methods are not scalable since their inference requires reversing a $\mathcal{O}(N^2)$-dimensional diffusion process. GDSS ignores the hierarchical interpretation between nodes and edges by modeling node and edge representations in parallel. 


In observation of those limitations, we propose NVDiff, a novel graph generative model built on latent score-based generative models (LSGM)~\citep{vahdat2021score}. The generative process first samples a latent representation of the graph and then decodes nodes and edges conditioning on the latent representation. In principle, the dimension of the latent space is arbitrary. To guarantee node exchangeability, we propose a diffusion process on latent node vectors, thus significantly reducing the dimension of the diffusion process from $\mathcal{O}(N^2)$ to $\mathcal{O}(N)$ and boosting sampling speed. Built on the NVDiff framework, we introduce an attention-based score network capturing local and global contexts within the graph by leveraging the self-attention mechanism with a contextual vector. NVDiff optimizes the variational lower bound of log-likelihood, which can be optimized end-to-end. Our empirical study demonstrates that NVDiff outperforms previous methods over various graph generation tasks. By diffusing only the node vectors, NVDiff can learn to generate larger graphs with high quality in one-shot. 
\section{Background: Score-based Generative Models}\label{sec:sgm}

Denote a continuous time diffusion process as $(\mathbf{Z}^t)_{t\in[0,1]}$ with base distribution $q_0(\mathbf{Z}^0)$. In practice, $q_0$ can either be the data distribution or latent variable distribution. Assuming the diffusion process follows an It\^{o} SDE:
\begin{align}
    \mbox{d}\bZ^t = f(t)\bZ^t \mbox{d}t + g(t) \mbox{d}\bW^t, \label{sde}
\end{align}
where $\bW^t$ is the standard Wiener process. $f(t)$ and $g(t)$ are scalar functions denoting drift and diffusion coefficients, which are chosen such that $q_1(\bZ^1)=\mathcal{N}(\bZ^1;\mathbf{0},\bI)$. The forward SDE gradually corrupt the base distribution to a random noise distribution that is easy to sample from, e.g., Gaussian distribution. To sample from the base distribution, we first draw $\bZ^1$ from $\mathcal{N}(\bZ^1;\mathbf{0},\bI)$ and run a reverse-time SDE:
\begin{align}
    \mbox{d}\bZ^t = [f(t)\bZ^t -g(t)^2\nabla_{\bZ^t}\log{q_t(\bZ^t)}]\mbox{d}t + g(t) \mbox{d}\bar{\bW}^t, \label{reverse-sde}
\end{align}
where $\bar{\bW}^t$ is the reverse-time standard Wiener process, and $q_t(\bZ^t)$ is the marginal distribution of the forward SDE at time $t$. Since $q_t(\bZ^t)$ is analytically intractable, we use a noise conditional score network $\epsilon_{\theta}(\bZ^t,t)$ to fit the score function as:
\begin{align}
    \min_\theta~~\mathbb{E}_{t\sim U[0,1]}\Big[\frac{g(t)^2}{2}\mathbb{E}_{q_0(\bZ^0)}&\mathbb{E}_{q(\bZ^t|\bZ^0)}[\|\nabla_{\bZ^t}\log{q_t(\bZ^t)}\nonumber\\
    &-\epsilon_{\theta}(\bZ^t,t)\|^2_2]\Big],
\end{align}
where $q(\bZ^t|\bZ^0)$ denotes the transition kernel of forward SDE. Replacing $\nabla_{\bZ^t}\log{q_t(\bZ^t)}$ with the score network $\epsilon_{\theta}(\bZ^t,t)$ in Eqn.~(\ref{reverse-sde}) derives a score-based generative model (SGM). We denote the marginal distribution of SGM at $t=0$ as $p_\theta(\bZ)$.
    \label{eq:general-elbo}

\section{NVDiff: Graph Generation through Node Variable Diffusion} \label{sec:nvdiff}


\begin{figure*}
    \centering
    \includegraphics[width=0.8\textwidth]{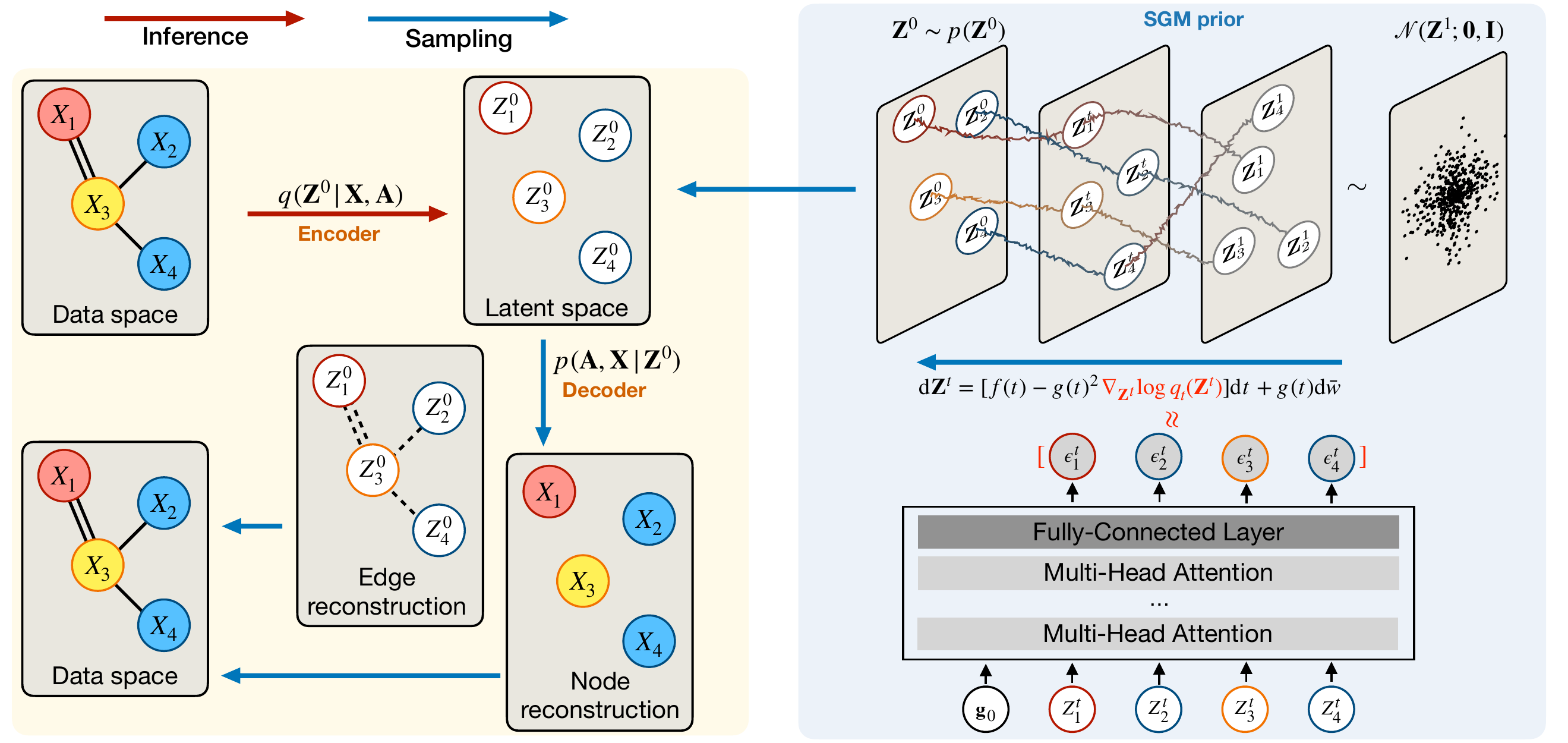}
\caption{In NVDiff, a graph $(\bA,\bX)$ is mapped to a set of latent node vectors $\bZ^0$ through an encoder, the joint distribution of the latent node vectors is model by a diffusion prior. To generate a graph, NVDiff first samples $\bZ^0$ by reversing the diffusion process, then generates nodes and edges by decoding the latent representation.}
\label{fig:framework}

\end{figure*}
\subsection{Graphs as Exchangeable Adjacency Matrices}

We consider an attributed graph $G$ with $N$ nodes and denote it by $(\bA, \bX)$. Here $\mathbf{A}\in\mathbb{R}^{N\times N\times K^e}$ denotes edge features, $\mathbf{X}\in\mathbb{R}^{N\times K^v}$ denotes node features, and feature dimensions are $K^e$ and $K^v$.
We mainly consider the case where nodes and edges are categorical data with one-hot vector representation. 

An exchangeable generative model $p$ requires $p(\mathbf{A},\mathbf{X})\stackrel{d}{=}p(\mathbf{A}[\pi,\pi,:],\mathbf{X}[\pi,:])$ for any permutation $\pi:[N]\rightarrow[N]$. $\mathbf{A}[\pi,\pi,:]$ represents the resulting matrix by jointly permuting the first and the second coordinates of $\mathbf{A}$, and $\mathbf{X}[\pi, :]$ represents permuting rows of $\mathbf{X}$ with $\pi$. With proper model design in the variational auto-encoder framework (VAE) by introducing intermediate latent variables, we can build flexible graph generative models without sacrificing exchangeability.




\subsection{The VAE Framework}
\label{form}
Variational Graph Auto-Encoder (VGAE)~\cite{kipf2016variational} is a variational auto-encoder that models the distribution of adjacency matrices. It imposes a Gaussian prior $p(\bZ)$ over node vectors. During training, it maps $(\bA, \bX)$ to $\bZ$ using an encoder $q_{\phi}(\bZ | \bA , \bX)$ and then recover $\bA$ using a decoder $p_\psi(\bA | \bZ)$. 


Our work implements the VGAE framework and makes several important differences. (1) We use SGM as our prior $p_\theta(\bZ)$ to increase expressiveness. (2) We use decoder $p_\psi(\bA,\bX|\bZ)$ to model both nodes and edges. (3) We make adjustment to the encoder $q_\phi(\bZ | \bA, \bX)$ to facilitate model training. We highlight our framework in Figure \ref{fig:framework}.

Learning VAE requires maximizing the variational lower bound $\calL(\phi, \psi, \theta)$ of log-likelihood $\log{p(\bA,\bX)}$ as
\begin{align}
\max_{\phi, \psi, \theta}~&\calL(\phi, \psi, \theta) \!= \mathbb{E}_{q_\phi(\bZ|\bA,\bX)}\big[\log{p_{\psi}(\bA,\bX|\bZ)}\big] \!\nonumber\\
&+\! \mathbb{E}_{q_\phi(\bZ|\bA,\bX)} \big[\log p_\theta(\bZ)\big]\!+\!\mathbb{H}\big[q_\phi(\bZ|\bA,\bX)\big]. \label{vlb}
\end{align}

In principle, $\bZ$ can be a latent variable with any dimension. Moderately compressed $\bZ$ enjoys compact representation and improves sampling speed of SGM. In NVDiff, we choose $\bZ \in \bbR^{N \times d}$, where each row corresponds to a latent node vector. To generate graphs, we first draw node vectors $\bZ\sim p_\theta(\bZ)$ and draw $\bA,\bX\sim p_{\psi}(\bA,\bX|\bZ)$. Edge information is encoded in mutually-dependent node vectors through a learnable SGM prior $p_\theta(\bZ)$.

\subsection{Attention-based Score Network with Contextual Embedding} \label{subsec:ncsn}




In NVDiff, learning $p_\theta(\bZ)$ is equivalent to learning the score network $\epsilon_\theta(\bZ^{t}, t):\mathbb{R}^{N\times d}\times \mathbb{R}\rightarrow \mathbb{R}^{N\times d}$. To guarantee exchangeability of the SGM prior, $\epsilon_\theta(\bZ^{t}, t)$ needs to be permutation-equivariant in rows, i.e., $\epsilon_\theta(\bZ^t, t)[\pi,:] = \epsilon_\theta(\bZ^t[\pi,:], t)$ for any permutation $\pi$~\citep{niu2020permutation}.


To guarantee permutation-equivariance without losing expressiveness, we use multi-head self-attention layers (MHA) to construct $\epsilon_\theta(\bZ^{t},t)$. Following~\citet{song2020score}, we transform $t$ to its positional encoding vector $\bt = \mathrm{pe}(t)$ as an input to the initial  mutli-layer perceptron (MLP) layer:
\begin{align}
    \bH_{0} = \mathrm{mlp}^{\mathrm{in}}([\bZ^t; \bone \bt^\top]),
\end{align}
where $[\cdot; \cdot]$ represents stacking two matrices by aligning their rows, and the $\mathrm{mlp}^\mathrm{in}(\cdot)$ applies to rows of its argument. 


Then we stack $L$ MHA layers to process the input $\bH_0$ and incorporate a learnable global vector $\bg_{0}$ to carry graph information through MHA layers:
\begin{align}
    (\bg'_{\ell}, \bH_{\ell}) = \mathrm{mha}([\bg_{\ell-1}; \bH_{\ell-1}]),\nonumber\\ 
    \bg_{\ell} = \bg_{\ell-1} + \bg'_{\ell}, 
    \quad
    \ell = 1,\ldots, L.
\end{align}

Finally, we predict the score from the representation $\bH_{L}$:
\begin{align}
    \hat{\bepsilon} = \mathrm{mlp}^{\mathrm{out}}(\bH_{L}).
\end{align}

{The input of the score network $\bZ^t$ is sampled from a transition distribution $q(\bZ^t|\bZ^0)=\mathcal{N}(\bZ^t;\bmu_t(\bZ^0),\sigma_t\bI)$, which has a closed form with $\bmu_t(\cdot)$ and $\sigma_t$ determined by $t$. Since $\bZ^t$ naturally carries blurred structural information of each node, we use a contextual vector $\bg_0$ to carry the global information by attending all the node vectors at the first MHA layer.} The following MHA layers gradually mix information between the contextual vector $\bg$ and the node vectors $\bH$ through self-attention until getting the output score.


To draw a sample $\bZ^0$ from $p_\theta(\bZ)$, we first draw $\bZ^1$ from $\mathcal{N}(\bZ^1;\bzero, \bI)$ and run the reverse-time SDE (Eqn.~(\ref{reverse-sde})) with true score function $\nabla_{\bZ^t}\log{q_t(\bZ^t)}$ replaced by the score network $\epsilon_\theta(\bZ^{t},t)$~\citep{song2020score}.

\subsection{Graph Decoding}
The decoder predicts node attributes, edges, and edge attributes from node representations $\bZ$. 
We first transform $\bZ$ to get the node representation $\bS \in \bbR^{N \times d'}$ and the edge representation $\bR \in \bbR^{N \times N \times d''}$ from the penultimate layer of the decoder.
\begin{align}
(\bS, \bR) = \mathrm{gnn}_{\psi}(\bZ, \bE), ~ \bE = [(\bz_i - \bz_j)^2: i, j = 1, \ldots, N].
\end{align}
Here $\bE\in\bbR^{N\times N\times d}$ is a tensor, whose $(i,j)$-th vector slice denotes the element-wise squared difference between two node vectors $\bz_i$ and $\bz_j$.
$\mathrm{gnn}_\psi(\cdot,\cdot)$ is a $M$-layer graph neural network that computes on a fully-connected graph with $\bZ,\bE$ being the node and edge inputs. The $\ell$-th layer of the GNN updates nodes and edges from $(\bS^{\ell-1},\bR^{\ell-1})$ to $(\bS^{\ell},\bR^{\ell})$ as follow: 
\begin{align}
    &\br_{i,j}^{\ell}= \mathrm{mlp}^e([\bW^{e}\bs_{i}^{\ell-1} + \bW^e\bs_{j}^{\ell-1};\br_{i,j}^{\ell-1}]),\nonumber\\
    &\bs_{i}^{\ell}= \mathrm{mlp}^v\Big(\Big[\sum_{j:j\neq i}^{N}\bW^{v}\br_{i,j}^{\ell};\bs_{i}^{\ell-1}\Big]\Big).
\end{align}
Here $\bs_i^\ell$ is the $i$-th row in $\bS^\ell$, $\br_{i,j}^\ell$ is the $(i,j)$-th vector slice in $\bR^\ell$,  we denote $(\bS^0,\bR^0):=(\bZ,\bE)$ as inputs to the GNN, and $(\bS,\bR):=(\bS^M,\bR^M)$. We recover node features and edge features respectively:
\begin{align}
p(\bA, \bX | \bZ) = p(\bA | \bR) p(\bX | \bS) = \prod_{i = 1}^{N} p(\bx_i | \bs_i) \prod_{j < i}  p(\ba_{i,j} | \br_{i,j}) ,
\end{align}
where $\bx_i$ denotes the $i$-th row in $\bX$ and $\ba_{i,j}$ denotes the $(i,j)$-th vector slice in $\bA$. Since the adjacency matrix is symmetric, we consider the lower triangle only.   

Now we are ready to recover individual features for edges and nodes. When we consider an edge feature, we treat a non-edge ($\ba_{i,j} = \bzero$) separately from an actual edge label. 
We specify $p(\ba_{i,j} | \br_{i,j})$ with two set of probabilities:  
\begin{align} 
&p(\ba_{i,j} = \bzero | \br_{i,j}) = \mathrm{mlp}^b_{\psi}(\br_{i,j}), \\
&p(\ba_{i,j}  | \ba_{i,j} \neq \bzero,  \br_{i,j}) = \mathrm{Categorical}(\ba_{i,j}; \mathrm{mlp}^e_{\psi}(\br_{i,j})).
\nonumber\end{align}
Here $\mathrm{mlp}^b_{\psi}(\cdot)$ calculates the Bernoulli probability of the non-existence of edge $(i,j)$, and $\mathrm{mlp}^e_{\psi}(\cdot)$ calculates the categorical probabilities for the label of edge $(i,j)$. 

Then we consider node features and define:
\begin{align}
p(\bx_i | \bs_i) = \mathrm{Categorical}(\bx_i; \mathrm{mlp}^n_{\psi}(\bs_i)).
\end{align}
Here $\mathrm{mlp}^n_{\psi}(\cdot)$ is applied to get the probabilities of the categorical distribution. We use sigmoid activation function in the last layer of $\mathrm{mlp}^b_\psi(\cdot)$ and softmax activation function in the last layer of $\mathrm{mlp}^e_\psi(\cdot)$ and $\mathrm{mlp}^n_\psi(\cdot)$.

\subsection{Graph Encoding}
We use a mean-field distribution as   the encoder and parameterize it with a GNN:
\begin{align}
    q_\phi(\bZ|\bA,\bX) = \calN(\bZ; \bM, \sigma^2 \bI), ~ \bM = \mathrm{gnn}_\phi(\bA, \bX).
\end{align}
Here the encoder is a Gaussian distribution, whose mean is computed from a GNN. The architecture details are described in Appendix. To simplify the optimization objective, we fix the variance of the encoder and only optimize its mean parameters. 
By doing so, we constrain flexibility of the encoder and encourage the SGM prior to match the aggregated posterior.
When we apply a GNN to $(\bA, \bX)$, we treat non-edges as an extra edge-type and also pass messages through them. Following \citet{satorras2021n}, we also inject random noise to each node to break the symmetry of the graph such that nodes are separable. We find these measures improve the model performance. 

\subsection{Training}

In this section, we present the training procedure that iteratively optimizes the encoder, the decoder, and the diffusion prior. The training procedure is shown in Alg. \ref{alg}.  In the optimization objective $\calL(\phi,\psi,\theta)$ in Eqn.~(\ref{vlb}), the first term $\mathbb{E}_{q_\phi(\bZ|\bA,\bX)}\big[\log{p_{\psi}(\bA,\bX|\bZ)}\big]$ is the reconstruction term. We apply the reparameterization technique \citep{kingma2013auto} and use Monte Carlo samples to estimate the expectation.
\citet{vahdat2021score} shows that the cross-entropy term can be calculated through score matching:
\begin{align}
  \mathbb{E}_{q_\phi(\bZ|\bA,\bX)}& \big[\log p_\theta(\bZ)\big] \nonumber\\
  &=\bbE_{t,\bepsilon,q_\phi(\bZ^0,\bZ^t|\bA,\bX)}\Big[\frac{g(t)^2}{2}||\bepsilon
  -\bepsilon_\theta(\bZ^t,t)||^2_2\Big]
\end{align}
where $t\sim U[0,1]$, $\bepsilon\sim\mathcal{N}(\bepsilon;\bzero,\bI)$, and $q_\phi(\bZ^0,\bZ^t|\bA,\bX)=q(\bZ^t|\bZ^0)q_\phi(\bZ^0|\bA,\bX)$. Here $q(\bZ^t|\bZ^0)$ is a Gaussian transition kernel. The cross-entropy term is also estimated through Monte Carlo method. We describe the details of computing the cross-entropy term in Appendix.
The entropy term $\bbH[q_\phi(\bZ|\bA,\bX)]$ in Eqn. (\ref{vlb}) is a constant because we fix the variance of the posterior distribution.  



Recall that $\calL(\phi,\psi,\theta) \!=\! \mathbb{E}_{q_\phi(\bZ|\bA,\bX)}\big[\log{p_{\psi}(\bA,\bX|\bZ)}\big] \!+\! \operatorname{KL}(q_\phi(\bZ|\bA,\bX)||p_\theta(\bZ))$. We use KL annealing to dynamically balance optimization between the $\operatorname{KL}$ term and the rest objective functions during training. KL annealing effectively prevents the variational posterior from converging to a premature diffusion prior.

 Sampling novel and unique graphs requires the decoder to be robust  to noise in node vectors, since the diffusion prior may generate node vectors unseen during training. We introduce a consistency regularization term to encourage the decoder to generalize better, by adapting to a perturbed latent variable $\tilde{\bZ}$. We define:
\begin{align}
    &\calL_\mathrm{Reg}\!\!=\!\!\bbE_{\tilde{\bepsilon}, q_\phi(\bZ|\bA,\bX)}\big[\!\!-\!\log{p_\psi(\bA,\bX|\tilde{\bZ})}\big], \nonumber \\
    &\mbox{with}~ \tilde{\bepsilon}\!\sim\!\calN(\bzero,\sigma_1\bI),~\bZ\!\sim\! q_\phi(\bZ|\bA,\bX),~\tilde{\bZ}\!=\!\bZ+\tilde{\bepsilon}.
\end{align}
Here $\sigma_1$ is a hyperparameter. We replace the reconstruction term with $\calL_\mathrm{Reg}$ and finetune the decoder.

\begin{algorithm}[t]
  \caption{{Training Procedure for NVDiff}}
  \begin{algorithmic}
    \STATE {\bfseries Input:} Graph dataset $\mathcal{G}=\{(\bA_1,\bX_1),\ldots\}$, encoder $q_\phi$, encoder $p_\psi$, diffusion prior $p_\theta$, transition kernel $q$, diffusion coefficient $g(\cdot)$
      \FOR{$(\bA,\bX) \in \calG$}
        \STATE \textbf{[Optimize the encoder and the decoder]}
        \STATE ~~~~Draw $\bZ^0\sim q_\phi(\bZ|\bX,\bA)$, $t\sim U[0,1]$, $\bepsilon\sim\mathcal{N}(\bepsilon;\bzero,\bI)$, $\bZ^t\sim q(\bZ^t|\bZ^0)$.
        \STATE ~~~~Optimize $\mathcal{L}_\mathrm{VAE}=-\log{p_\psi(\bA,\bX|\bZ^0)}+\frac{ g(t)^2}{2}||\bepsilon-\bepsilon_\theta(\bZ^t,t)||^2_2$ over $\phi$ and $\psi$ with one step.
        \STATE \textbf{[Optimize the SGM prior]}
        \STATE ~~~~Optimize  $\mathcal{L}_\mathrm{SGM}=\frac{g(t)}{2}||\bepsilon-\bepsilon_\theta(\bZ^t,t)||^2_2$ over $\theta$ with one step.
      \ENDFOR
  \end{algorithmic}
  \label{alg}
\end{algorithm}

\begin{table*}[t]
    \centering
    \small
    \caption{Performance on molecule generation. Numbers in bold indicate that the methods are the best, and numbers underlined are the second best. Extended NSPDK, Uniqueness metric as well as standard deviations are in Appendix.}
    \begin{tabular}{lcccccc}\hline
         &  \multicolumn{3}{c}{QM9} & \multicolumn{3}{c}{ZINC250K}\\
         &  Validity$\uparrow$ & FCD$\downarrow$ & Time (s)$\downarrow$ &Validity$\uparrow$ & FCD$\downarrow$  & Time (s)$\downarrow$ \\\hline
         GraphDF & 82.67 & 10.816 &5.35e4 & 89.03& 34.202 & 6.03e4\\
         MoFlow & 91.36 & 4.467 & \textbf{4.60} &63.11 & 20.931 &\textbf{24.5}\\
         GraphCNF &95.00&\underline{1.629}&59.48&\underline{95.26}&\underline{14.786}&1.04e2\\
         EDP-GNN & 47.52  & 2.680 &4.40e3 & 82.97 & 16.737 & 9.09e3\\
         GDSS & \underline{95.72} & 2.900 &1.14e2&\textbf{97.01} & 14.656 & 2.02e3 \\\hline
         NVDiff & \textbf{95.79} & \textbf{1.131} &\underline{10.71} &85.63&\textbf{4.019} &\underline{90.2}\\\hline
    \end{tabular}
    \label{tab:molecule}
\end{table*}
\section{Related Works} \label{sec:related}
\noindent\textbf{Score-based generative models.} Score-based generative models~\citep{song2019generative,ho2020denoising,song2020score} have shown effective at the generation of data in various domains \citep{ho2020denoising,chen2020wavegrad,niu2020permutation}. In SGMs, data are first diffused with gradually increasing noise, and the model learns to estimate the score at each noise scale. During generation, an SGM first randomly draws some noise from a simple distribution and then decreases the noise scales sequentially to obtain a sample. Instead of generating data directly, \citet{mittal2021symbolic,sinha2021d2c,wehenkel2021diffusion,vahdat2021score} utilize SGM in latent space and turn it into a flexible prior by combining the VAE framework~\citep{kingma2013auto}. {Unlike approaches that use more complex prior \citep{dai2019diagnosing,klushyn2019learning}, SGM prior addresses the posterior mismatch problem by construction \citep{sinha2021d2c}. Combining SGM prior into VAE framework also enables faster sampling as well as hierarchically generating data.}

\noindent\textbf{Graph generation models.} Current graph generative models sample graphs either autoregressively or in one-shot. 
Autoregressive graph generative models generate graphs by adding nodes and edges sequentially~\citep{you2018graphrnn,li2018multi,li2018learning,jin2018junction,liao2019efficient,shi2020graphaf,zang2020moflow,bacciu2020edge,dai2020scalable,goyal2020graphgen,tran2020deepnc,luo2021graphdf, bengio2021flow}. 
Training such models requires a particular node ordering from the graph to pin down the generation sequence, thus unable to preserve the model exchangeability~\citep{chen2021order}. Besides, the autoregressive models usually fail to learn the global structure of a graph as they can not capture the long-term dependency. 
The other type of models generate graph in one-shot~\citep{simonovsky2018graphvae,liu2018constrained,madhawa2019graphnvp,kajino2019molecular,bradshaw2019model,grover2019graphite,lim2020scaffold,jin2020hierarchical,guo2020node,lippe2020categorical,samanta2020nevae,niu2020permutation,garcia2021n,jo2022score}. Unlike autoregressive models, exchangeability can be preserved through carefully designing the model architecture. However, those models need to generate nodes and edges in parallel and can not scale up to large graphs. 
On par with the one-shot graph generative models, VGAEs~\citep{kipf2016variational,mehta2019stochastic,liu2019graph,li2020dirichlet} adopt the variational inference framework to learning node representations that are useful for downstream tasks, such as unsupervised learning or node clustering. However, The merits of using such approaches in graph generation, such as the exchangeable property and the parallel sampling scheme, have not been fully exploited. 
Our work builds on the VGAE framework and utilizes a learnable prior based on SGM.

\section{Experiments} \label{sec:exp}

\begin{table*}[t]
    \centering
    \small
    \caption{Performance on generic graph generation. Numbers in bold indicate that the methods are the best, numbers underlined are the second best. ``OOM'' denotes out of memory.}
    \begin{tabular}{lcccccccc}\hline
    &\multicolumn{4}{c}{Community-small} & \multicolumn{4}{c}{Ego-small}\\
    &Deg.$\downarrow$ & Clus.$\downarrow$ &Orbit$\downarrow$ &{NN$\downarrow$} & Deg.$\downarrow$& Clus.$\downarrow$ &Orbit$\downarrow$&{NN$\downarrow$}\\\hline
    GraphRNN & 0.080 & 0.120 & 0.040 &{0.997}& 0.090 & 0.220 & \underline{0.003} &{\textbf{1.094}}\\
    GRAN& 0.070 & \underline{0.045} & 0.021 &\underline{0.701}& 0.020&0.126&0.010 &{\textbf{1.095}}\\
    GraphCNF & \textbf{0.021} & 0.141 & 0.044 &{0.876}& \underline{0.011} & \textbf{0.011} & \textbf{0.001} &{\textbf{1.094}}\\
    EDP-GNN & 0.053 & 0.144 & 0.026 &{0.789}& 0.052 & 0.093 & 0.007&{\textbf{1.095}}\\
    GDSS &  \underline{0.045} & 0.086 & \textbf{0.007} &{0.858}& 0.021 & \underline{0.024} & 0.007 &\underline{1.097}\\\hline
    NVDiff & \textbf{0.021} & \textbf{0.035} & \underline{0.018} &{\textbf{0.688}}& \textbf{0.005} & 0.045 & \textbf{0.001} &{\textbf{1.092}}\\\hline
    &\multicolumn{4}{c}{Community} & \multicolumn{4}{c}{Ego}\\
    &Deg.$\downarrow$ & Clus.$\downarrow$ &Orbit$\downarrow$ &{NN$\downarrow$} & Deg.$\downarrow$& Clus.$\downarrow$ &Orbit$\downarrow$&{NN$\downarrow$}\\\hline
    GraphRNN & \textbf{0.014} &\textbf{0.002}& \underline{0.039} &{1.452}& 0.077 &0.316& \underline{0.030} &\underline{0.192}\\
    GRAN &  \underline{0.085} & \underline{0.068} &0.042 &{1.406}& \underline{0.064} & \underline{0.279} & 0.039 &{0.196}\\
    GraphCNF & 0.707 & 1.305 & 0.194 &{1.356}& OOM&OOM&OOM &{OOM}\\
    EDP-GNN & 0.137 &1.104 & 0.051 &\underline{1.146}& 0.069&0.447&0.055 &{\textbf{0.189}}\\
    {GDSS} & {0.099} & {0.327} & {0.162} &{1.203}&{0.134} & {0.372} & {0.139} &{0.295}\\\hline
    NVDiff &0.093&0.086&\textbf{0.021}&{\textbf{0.982}}& \textbf{0.045}& \textbf{0.091}&  \textbf{0.004} &{0.195}\\\hline
    \end{tabular}
    \label{tab:generic}
\end{table*}

\begin{figure*}
    \centering
    \includegraphics[width=\textwidth]{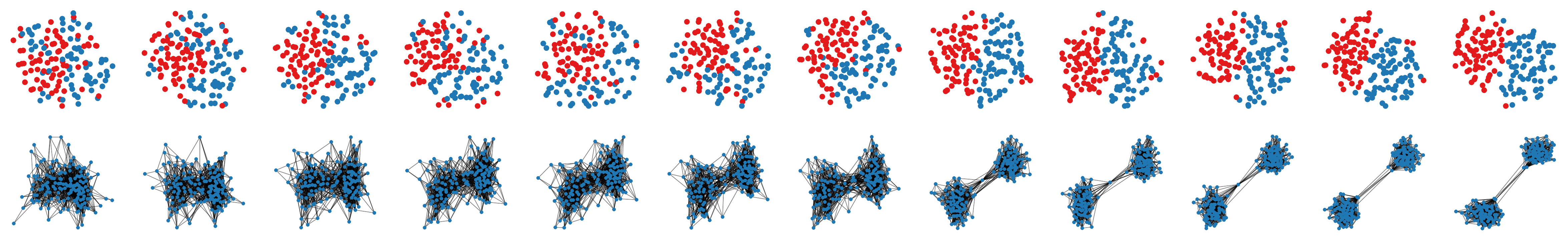}
    \caption{Visualization of the denoising dynamics on Community dataset ({evolve from left to right}). Top row: t-SNE visualization over the node vectors, with color indicating the community membership. Bottom row: graphs generated from the corresponding node vectors via the decoder.}
    \label{fig:dynamic}
\end{figure*}
{We design a set of experiments to demonstrate that NVDiff can generate high-quality graphs with latent node vectors. Specifically, in Section \ref{sec:exp-quan} we quantitatively evaluate the sample quality and sampling speed for selected baseline models and NVDiff over various metrics. In Section \ref{sec:exp-qual} we investigate NVDiff from the perspective of the generation dynamics, effectiveness of the contextual embedding, and graph visualization. More experiment results can be found in the Appendix.}
\subsection{Experiment Setup}
\label{sec:exp-setup}
Here we briefly discuss the datasets, baselines, and evaluation metrics we use in the experiments. Further details can be found in Appendix.

\noindent\textbf{Datasets.} We use six datasets, including two molecular datasets and four generic graph datasets. For molecular datasets, we consider QM9 ~\citep{ramakrishnan2014quantum} and {ZINC250K} ~\citep{irwin2012zinc}. For generic graph datasets, we consider {Community} and {Ego} and the small versions of them.


\noindent\textbf{Baselines.} We consider the following baselines: GraphRNN and GRAN ~\citep{you2018graphrnn,liao2019efficient} are autoregressive models. {GraphDF}~\citep{luo2021graphdf} is an autoregressive flow-based model. Moflow, GraphCNF \citep{zang2020moflow, lippe2020categorical} are flow-based models that generate graphs in one-shot. EDP-GNN and GDSS ~\citep{niu2020permutation,jo2022score} are one-shot score-based graph generative models. For molecule generation, we compare NVDiff with GraphDF, Moflow, GraphCNF, EDP-GNN, and GDSS. We compare generic graph generation with GraphRNN, GRAN, GraphCNF, EDP-GNN, and GDSS.

\noindent\textbf{Evaluation metrics.} We evaluate the generated molecules with the following metrics: {Validity} is the fraction of generated molecules with valid chemical valency. And {Uniqueness} is the fraction of unique molecules. 
We evaluate the {Maximum Mean Discrepancy (MMD)} ~\citep{gretton2012kernel} between the test molecules and the generated molecules in terms of the {Neighborhood subgraph pairwise distance kernel (NSPDK)} ~\citep{costa2010fast}, which considers both structure and labels of the graph. 
{Fr\'echet ChemNet Distance (FCD)} ~\citep{preuer2018frechet} evaluates the diversity and the quality of the validly generated molecules by measuring the distance between feature vectors calculated for real and generated molecules. The FCD is arguably the most important metric since it captures the similarity between real and generated molecules' chemical properties. 
{For the generated generic graphs, we evaluate the MMD distance between the test graphs and the generated graphs in terms of four type of graph features, including the statistics of degrees (Deg.), clustering coefficients (Clus.), and orbit counts (Orbit) from \citet{you2018graphrnn}, and the neural-based graph embedding (NN) proposed by \citet{thompson2022evaluation}.}
We report the metrics of the molecule and generic graph generation using 10000 and 128 samples, respectively.


\subsection{Quantitative Analysis}
\label{sec:exp-quan}
Here we quantitatively estimate the quality of the generated molecules and generic graphs and report them in Table \ref{tab:molecule} and Table \ref{tab:generic}, respectively. {We also analyze the sampling speed of NVDiff using the Ego dataset.}

\begin{figure}[h]
    \centering
    \includegraphics[width=0.38\textwidth]{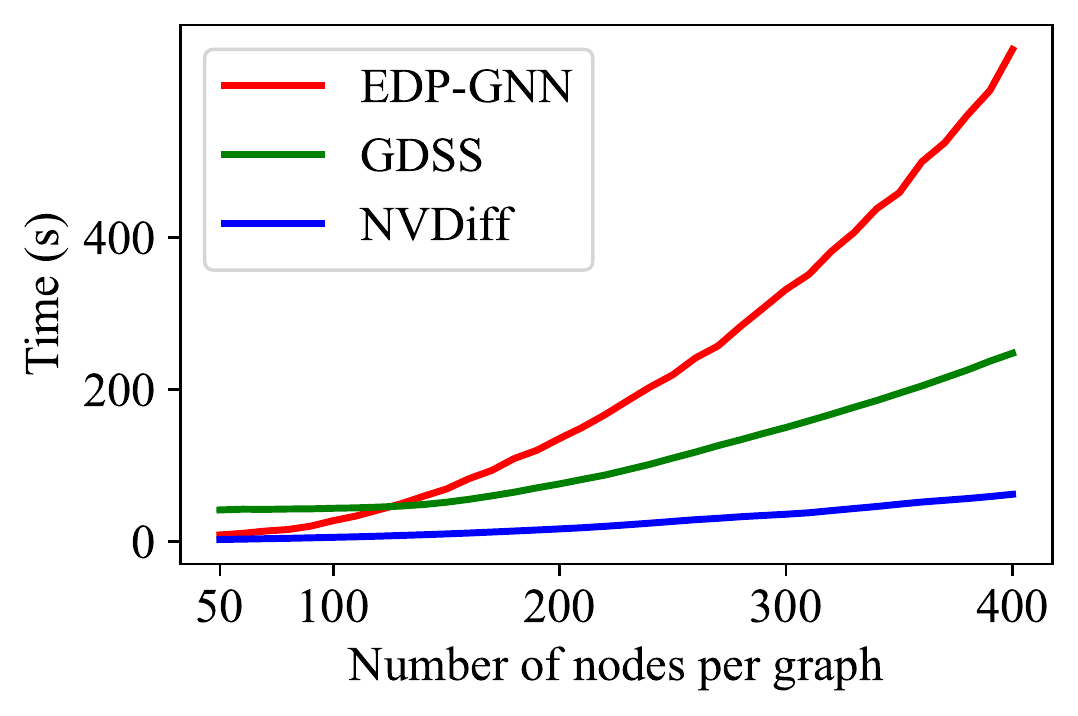} 
    \caption{{Sampling speed}}
    \label{tab:speed}
\end{figure}
\begin{figure*}[t]
    \small
    \centering
    \begin{tabular}{cccccc}
        &Reference &NVDiff & GraphDF & Moflow &  GraphCNF\\
        \rotatebox{90}{QM9}& 
        {\includegraphics[width=0.09\textwidth]{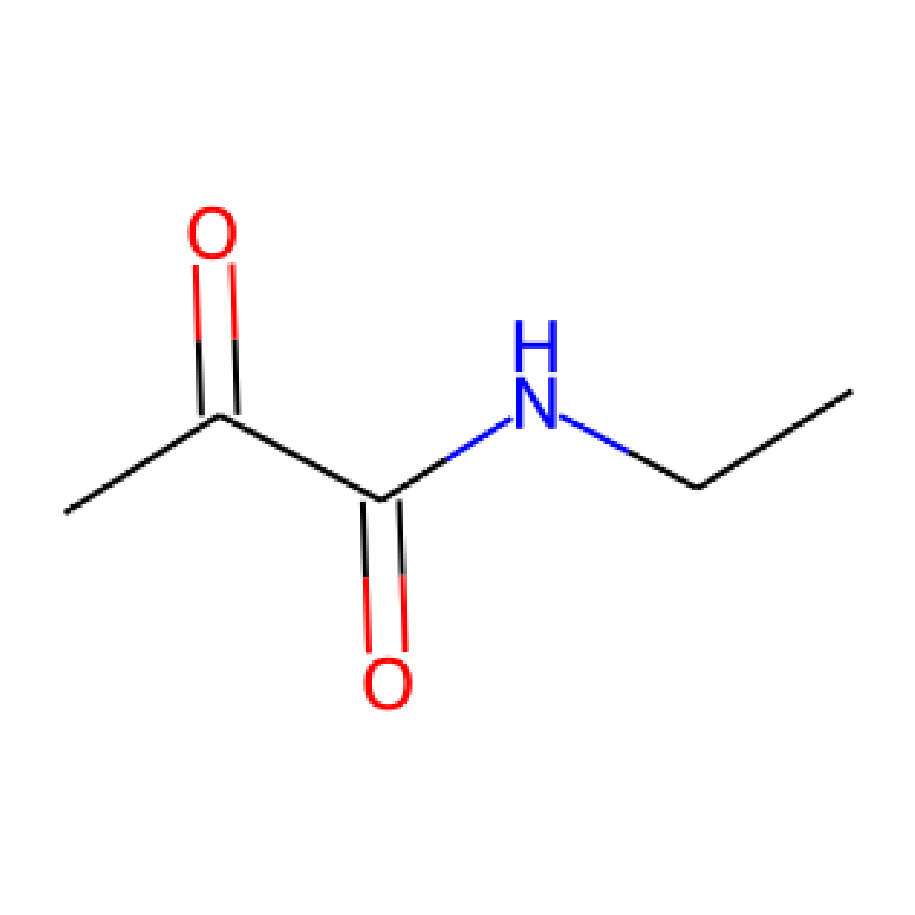}}& 
        {\includegraphics[width=0.17\textwidth]{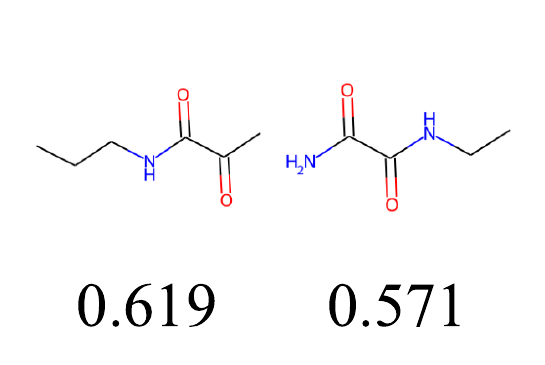}}&
        {\includegraphics[width=0.17\textwidth]{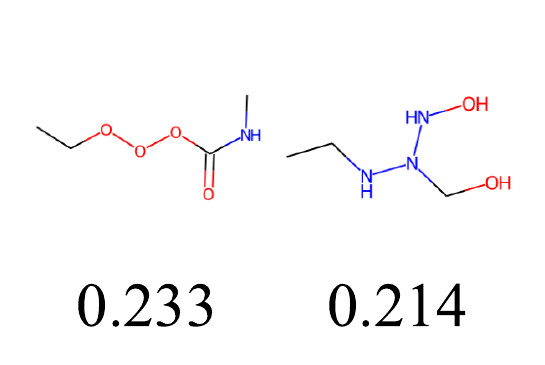}}&
        {\includegraphics[width=0.17\textwidth]{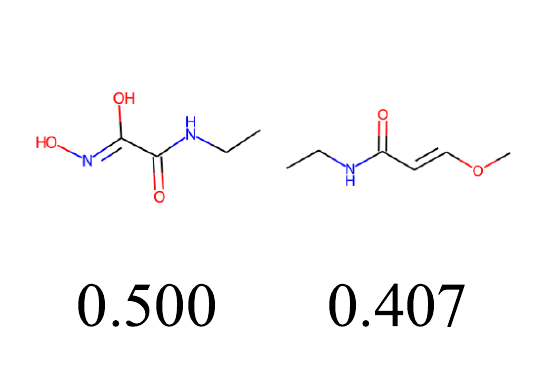}}&
        {\includegraphics[width=0.17\textwidth]{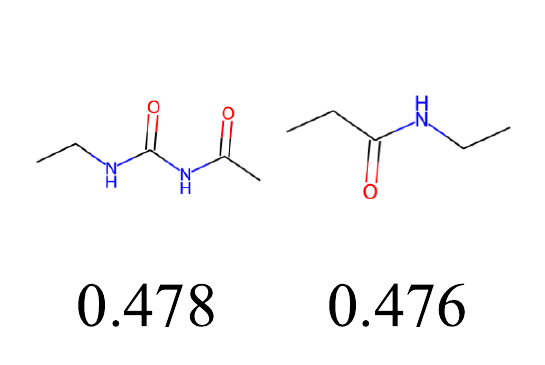}}\\
      
     &  {\includegraphics[width=0.09\textwidth]{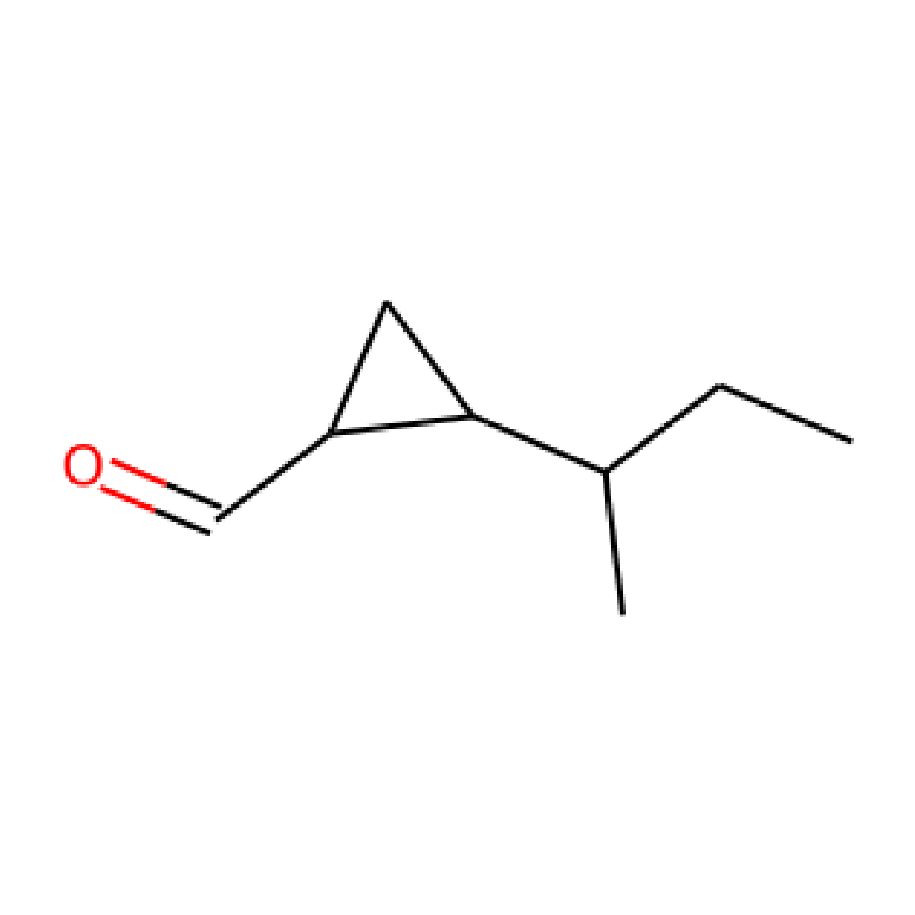}}& 
        {\includegraphics[width=0.17\textwidth]{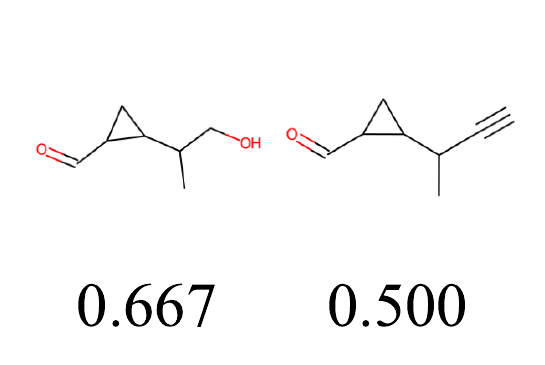}}&
        {\includegraphics[width=0.17\textwidth]{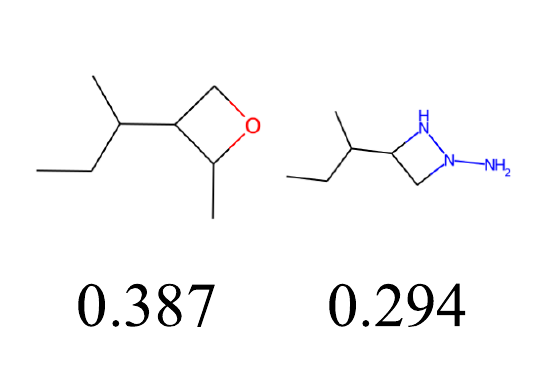}}&
        {\includegraphics[width=0.17\textwidth]{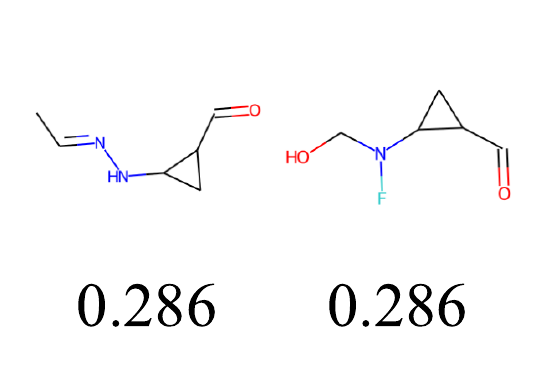}}&
        {\includegraphics[width=0.17\textwidth]{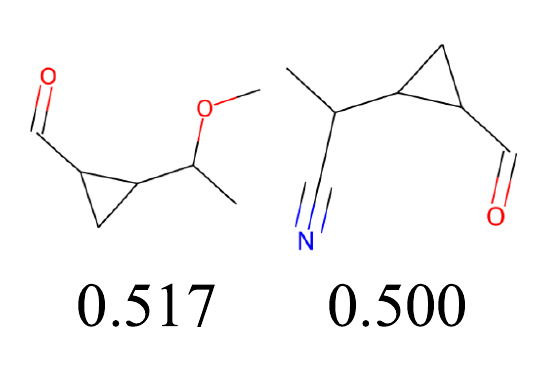}}\\
        \rotatebox{90}{ZINC250k}& {\includegraphics[width=0.09\textwidth]{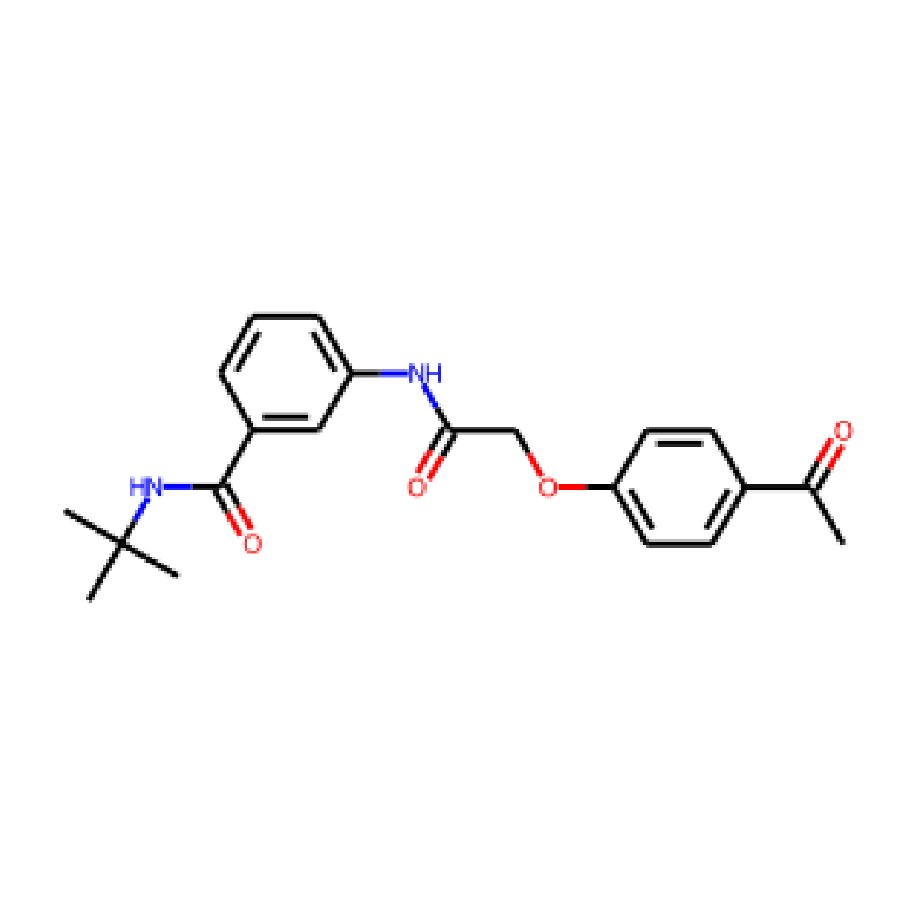}}&
      {\includegraphics[width=0.17\textwidth]{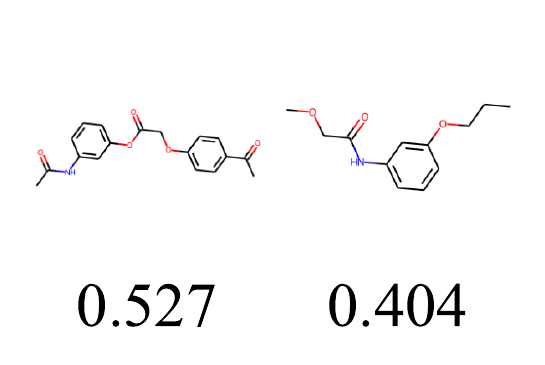}}&
     {\includegraphics[width=0.17\textwidth]{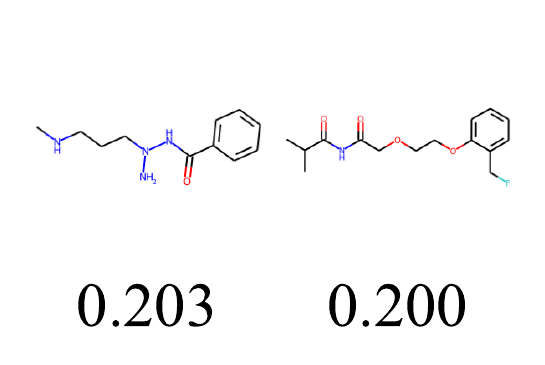}}&
      {\includegraphics[width=0.17\textwidth]{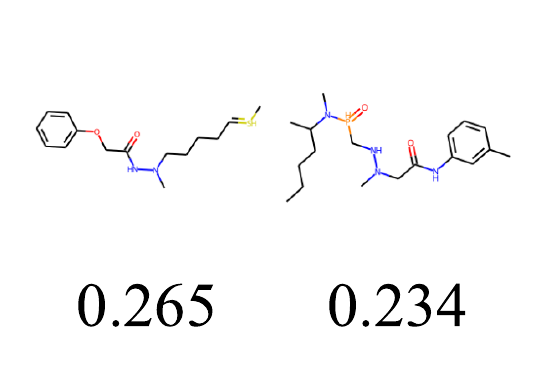}}&
      {\includegraphics[width=0.17\textwidth]{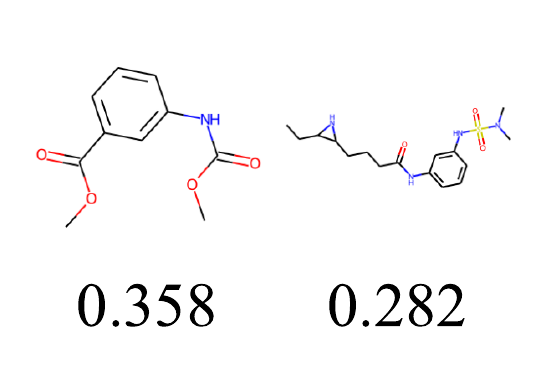}}\\
     &{\includegraphics[width=0.09\textwidth]{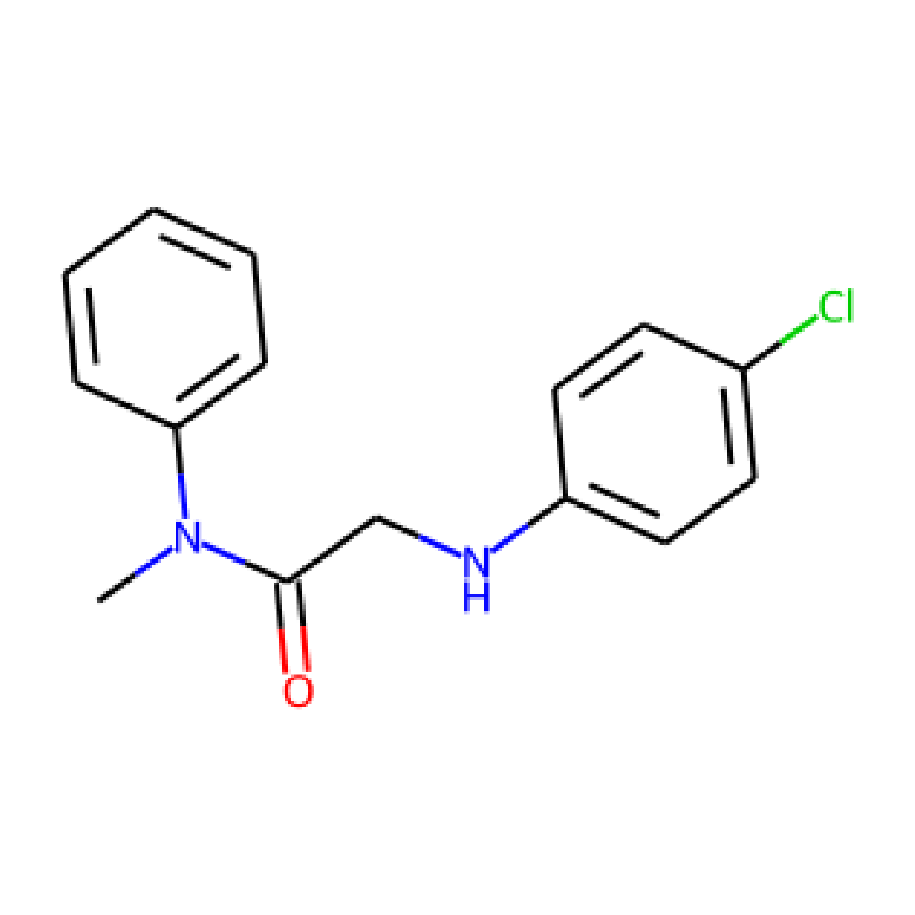}}& 
      {\includegraphics[width=0.17\textwidth]{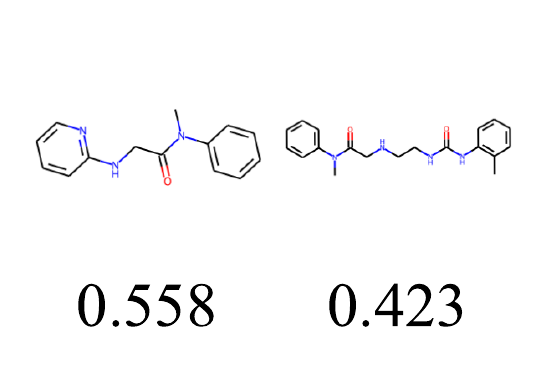}}&
     {\includegraphics[width=0.17\textwidth]{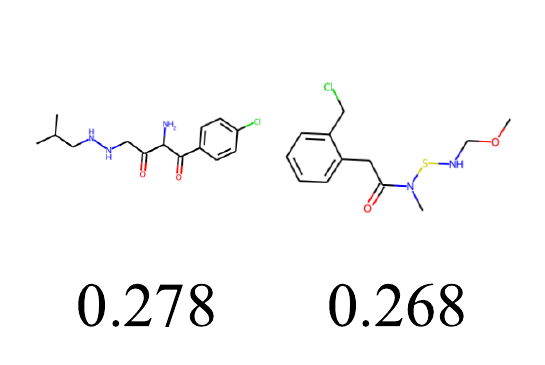}}&
      {\includegraphics[width=0.17\textwidth]{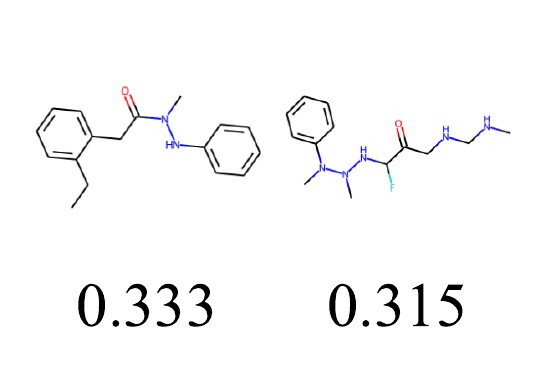}}&
      {\includegraphics[width=0.17\textwidth]{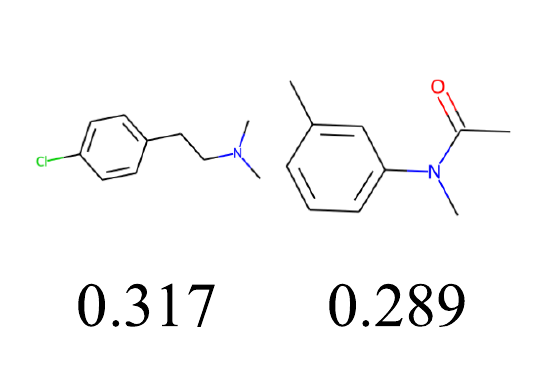}}
      \end{tabular}
      
    \caption{Molecule visualization with Tanimoto similarity. The left-most column visualizes the reference molecules. The top 2 molecules ranked by the similarity scores for each model are visualized. Extended results are in the Appendix.}
    \label{fig:mol_vis}
\end{figure*}
\noindent\textbf{Molecule generation.} NVDiff outperforms all baselines in terms of the FCD metric, thus superior in generating valid molecules that share highly similar characteristics with real molecules. However, on ZINC250K, we observe that NVDiff exhibits relatively lower validity. We argue that filtering invalid molecules is easy as it can be done by checking the valency rule while selecting realistic ones is difficult. We also observe that the sampling speed of NVDiff is much faster than most baselines.

\noindent\textbf{Generic graph generation.} We show the MMD evaluation on four datasets. Community and Ego contain massive nodes and thus are difficult to model by several baseline methods. We only report performance for baselines that converged in a reasonable time for these two datasets without consuming extensive computational resources. We observe that NVDiff outperforms all the baselines on most metrics. In GraphRNN, data are fit using BFS orderings, which are favored by datasets with community structure--the model naturally learns to generate communities sequentially. We argue that even without introducing such inductive bias, NVDiff still can capture the community pattern. In comparison, EDP-GNN and GraphCNF fail to capture the pattern of Community and Ego or exhibit poor performance. We hypothesize that the node-edge relationship becomes more complex as the graph size increases. So modeling nodes and edges in the diffusion prior is not a proper choice. Instead, the diffusion prior in NVDiff only needs to learn the node-to-node relationship, leaving the task of composing edges to the decoder.

\noindent\textbf{Sampling speed.} Here we compared against other score-based graph generative models in terms of the sampling speed. {We choose Ego for runtime analysis. Specifically, we manually fix the number of nodes the model will generate. For each setting, we sample 128 graphs and record the wall clock time. Figure \ref{tab:speed} shows the running time concerning the number of nodes of the generated graphs. NVDiff generates much faster than EDP-GNN and GDSS, especially over larger graphs.}

\begin{figure}[h]
    \centering
    \includegraphics[width=0.4\textwidth]{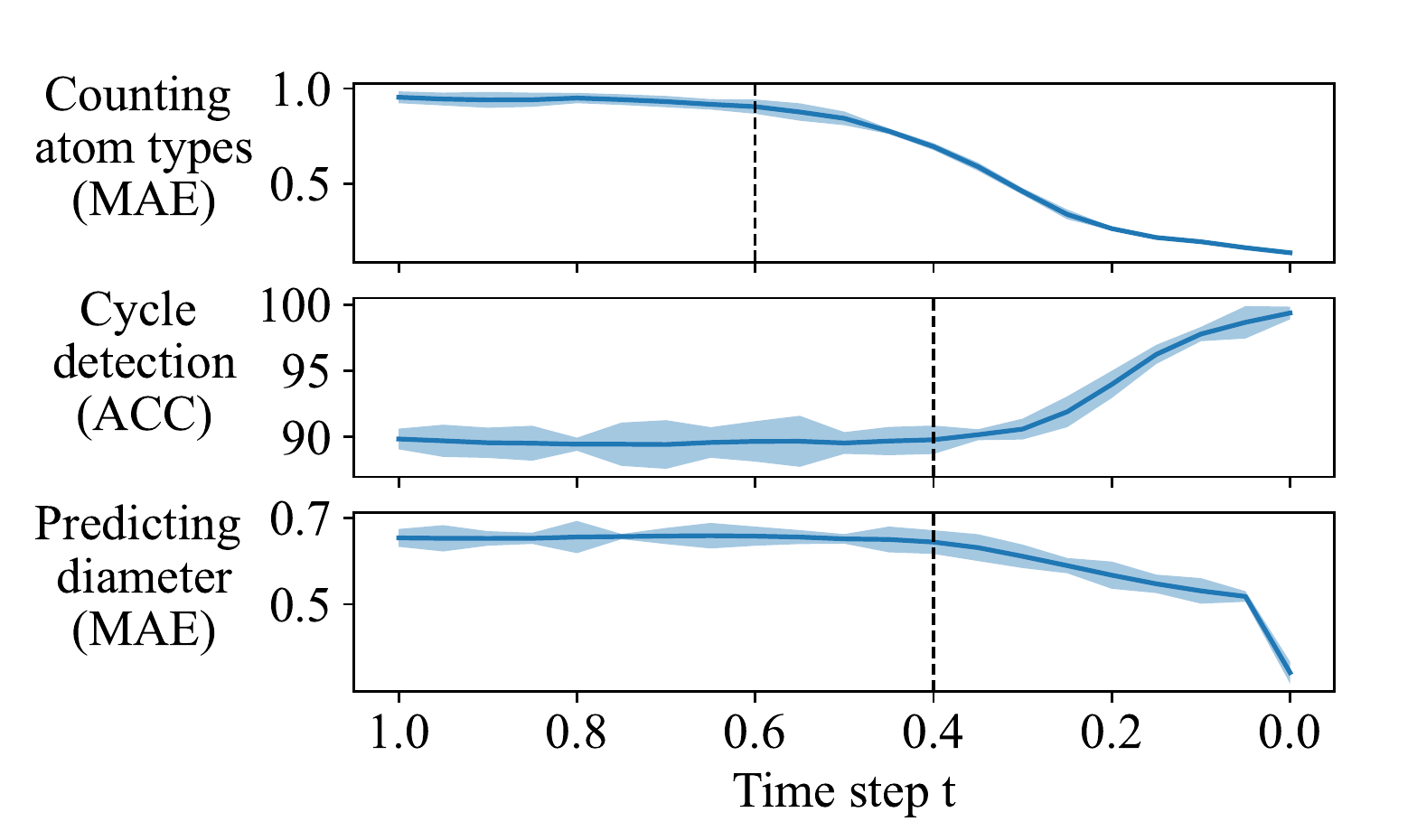}
    \caption{Downstream tasks with contextual vector}
    \label{fig:context}
\end{figure}

\subsection{Qualitative Analysis}
\label{sec:exp-qual}
We examine the effectiveness of NVDiff from three aspects: the denoising dynamics of node vectors, the quality of the generated molecules, and the graph statistics encoded in the contextual vector.

\noindent\textbf{Denoising dynamics of node vectors.}
Figure \ref{fig:dynamic} shows how NVDiff generates node vectors and forms a community graph progressively. We record the node vectors along the denoising step. For each time step, we use t-SNE~\citep{van2008visualizing} to compress the node vectors for visualization. We color each node with its community membership obtained from the spectrum clustering. For each intermediate node vectors, we generate the graph using the learned decoder.
As the denoising step increases, node vectors from different communities become separable.

\noindent\textbf{Sample quality.} We inspect the drug-likeness of the generated samples by measuring the Tanimoto similarity of the Morgan fingerprint ~\citep{bajusz2015tanimoto}. Specifically, we calculate the pair-wise Tanimoto similarity between each generated sample and a real molecule randomly chosen from the dataset. Given a reference molecule, we retrieve the top 2 generated molecules ranked by the similarity score and visualize them in Figure \ref{fig:mol_vis}. We analyze samples generated from NVDiff, GraphDF, Moflow, and GraphCNF. Compared to other baselines, NVDiff can generate molecules that share a large substructure with the actual molecules, illustrating a superior modeling capability.

\noindent\textbf{Graph statistics in contextual vector.} We investigate whether the introduced contextual vector contains meaningful information about the graph, which can be used for supervising the diffusion process. Here we focus on QM9. We hypothesize that the contextual vector contains both the node and structural information. To validate this, we design three downstream tasks: atom type counting, cycle detection, and diameter prediction. The first task focuses on the global node statistics, and the last two tasks focus on the local and global structural contexts, respectively. We report accuracy for the cycle detection task and Mean Absolute Error (MAE) for the others. Figure \ref{fig:context} demonstrates the downstream task performance of the contextual vectors extracted from different time steps. We can observe that all task performance improves incrementally as time step decreases (i.e., node vectors get less noisy). Moreover, the predictions are precise when $t$ is close to $0$, indicating that the contextual vector encodes significant contexts locally and globally about the graph.
Interestingly, we find that the performance of tasks related to node context starts to converge earlier than those related to structural context. One reasonable explanation is that the node context is formed earlier than the structural context, as the latter usually depends on the former. We highlight the time step when the task starts to converge in the figure with a dash line. 




\section{Conclusions and limitations} \label{sec:conclude}
This paper introduces NVDiff that generates novel and realistic graphs by taking the VGAE structure and uses SGM as its prior for latent node vectors. NVDiff guarantees exchangeability without losing expressiveness by implementing SGM with attention-based networks. When pairing with a properly designed GNN decoder, SGM can sample complex graphs in one shot and enjoys a fast sampling speed.

This paper has the following limitations: (1) NVDiff optimizes the variational lower bound of the likelihood, which corresponds to the KL divergence between the empirical distribution and the parameterized distribution. Optimization under other objectives is not discussed. (2) Current NVDiff does not support conditional graph generation, which can be used for optimizing molecular property.

\bibliography{references}
\newpage
\appendix
\setcounter{secnumdepth}{0}
\section{Implementation Details}
\subsection{Encoder}
Here we demonstrate the architecture of the encoder $\mathrm{gnn}_\phi(\cdot,\cdot)$. We treat non-edge as an extra edge type and use a GNN to output the encoding distribution. First, we create the input node representation $\bM^0$ for the encoder by injecting Gaussian Noise
\begin{align}
    \bM^0 = [\bX;\bepsilon^\mathrm{in}], \bepsilon^\mathrm{in}\sim\calN(\bepsilon^\mathrm{in};\bzero,\bI),
\end{align}
Denote $\mathbf{m}^\ell_i$ to be the $i$-th row of $\bM^\ell$, the $\ell$-th layer of the encoder updates the node representation as follow:
\begin{align}
    &\mathbf{m}^\ell_i=\mathbf{m}^{\ell-1}_i+\mathrm{mlp}^v([\mathbf{m}^\ell_i;\sum_{j\neq i}^N\mathbf{m}^\ell_{i,j}]),\nonumber\\ 
    &\mathbf{m}^\ell_{i,j} = \mathrm{mlp}^e([\ba_{i,j};\bW \mathbf{m}^{\ell-1}_{i}+\bW \mathbf{m}^{\ell-1}_{j}]).
\end{align}
Let $L'$ be the number of layer of the $\mathrm{gnn}_\phi(\cdot,\cdot)$. We denote $\bM:=\bM^{L'}$ being the mean of the encoding distribution.
\label{app:arch-enc}
\subsection{NVDiff-E}
To demonstrate the effectiveness of the introduced contextual vector, we propose a variant of the score network that diffuses both nodes and edges, which we call NVDiff-E. Compared to NVDiff, NVDiff-E exhibits a slower sampling speed and is incapable of modeling larger graphs. In molecule generation, where graphs are relatively small, NVDiff-E performs better than most methods, including NVDiff. Since NVDiff-E can operate directly on data space $(\bA,\bX)$, we only show how its score network is built. 

The score network for NVDiff-E $\epsilon^\mathrm{AX}_\theta(\bA^t,\bX^t, t)$ takes in the noisy inputs $(\bA^t,\bX^t)$ and time step $t\in [0, 1]$, and predict the scores for the edge and node features, respectively. We first inject time step information into the node and edge inputs:
\begin{align}
    &\bH^{e}_{0} = \mathrm{mlp}^{\mathrm{in},e}(\big[\bA^t;  \bone \bt^\top\big]),\nonumber\\
    &\bH^v_0 = \mathrm{mlp}^{\mathrm{in},v}(\big[\bX^t; \bone \bt^\top\big]),~\bt=\mathrm{pe}(t).
\end{align}

Here we define the edge-node attention layer~(ENA). We stack $L$ ENA layers to obtain the final representations $(\bH^e_L,\bH^{v}_L)$. We defer the details of the ENA module, and the following computations are repeated $L$ times:
\begin{align}
    (\bg_{\ell}, \bH^{v}_\ell,\bH^{e}_\ell) = \mathrm{ENA}(\bg_{\ell-1},\bH^{v}_{\ell-1},\bH^{e}_{\ell-1}),~ \ell=1,\ldots,L.
\end{align}
Here $\bg_\ell$ is the contextual embedding. We predict the scores from the final representations.
\begin{align}
    \hat{\bepsilon}^e = \mathrm{mlp}^{\mathrm{out},e}(\bH^{e}_{L}),~\hat{\bepsilon}^v = \mathrm{mlp}^{\mathrm{out},v}(\bH^{v}_{L})
\end{align}

\noindent\textbf{Edge-node attention.} Here we define ENA module. The ENA module is used to obtain ($\bg_\ell, \bH^{v}_\ell, \bH^{e}_\ell$) from $(\bg_{\ell-1}, \bH^{v}_{\ell-1}, \bH^{e}_{\ell-1})$:
\begin{align}
   &(\bg'_{\ell}, {\bH^{v}_\ell}') = \mathrm{mha}(\bg_{\ell-1},\bH^{v}_{\ell-1}),~\bg_\ell = \bg_{\ell-1}+\bg'_\ell, \nonumber\\
    &\bh_{i,\ell}^v={\bh_{i,\ell}^v}'+\sum_{j:j\neq i}\bW^v\bh_{i,j,\ell-1}^e, \nonumber\\
    &\bh_{i,j,l}^e = \mathrm{mlp}^e(\big[\bW^e\bh_{i,\ell}^v+\bW^e\bh_{j,\ell}^v;\bh_{i,j,\ell-1}^e\big]).
\end{align}
Here $\bh^v_{i,\ell}$ and ${\bh^v_{i,\ell}}'$ are the $i$-th row in $\bH^v_\ell$ and ${\bH^v_\ell}'$, respectively. And $\bh^e_{i,j,\ell}$ is the $(i,j)$-th vector slice in $\bH^e_{\ell}$. 

To draw a sample from NVDiff-E, we first sample $\bA^1$ and $\bX^1$ from $\calN(\bA^1;\bzero,\bone)$ and $\calN(\bX^1;\bzero,\bone)$, respectively, and run the reverse-time SDE to obtain $(\bA^0,\bX^0)$. Then we obtain $(\bA,\bX)$ by discretizing $(\bA^0,\bX^0)$:
\begin{align}
    &\bA = \mathrm{one\_hot}(\mathrm{argmax}(\bA^0)),\nonumber\\  
    &\bX = \mathrm{one\_hot}(\mathrm{argmax}(\bX^0)). 
\end{align}

\begin{table*}[t]
    \centering
    \caption{Hyperparameters of NVDiff used in the experiments.}
    \begin{tabular}{lcccccc}\hline
         &  QM9&ZINC250K&Community&Ego&Community-small&Ego-small\\\hline
        \noindent\textbf{Encoder}\\
        \# layers & 3 & 5 & 3 &3&3&3\\
        \# hidden dims &64&64&64&64&32&32\\
        \# latent dims &16&32&8&4&4&4\\
        \# noise dims &8&8&8&8&8&8\\
        fixed var. &0.01&0.0025&0.01&0.01&0.01&0.01\\
        \hline
        \noindent\textbf{Decoder} & \\
        \# layers & 3 & 3 & 1 &1&1&1\\
        \# hidden dims &64&64&64&64&32&32\\
        \# finetune epochs &200&200&0&0&0&0\\
        noise var. in finetuning  & 0.0001& 0.025&0&0&0&0\\
        \hline
        \noindent\textbf{SGM prior} & \\
        $\beta_0$&\multicolumn{6}{c}{0.1}\\
        $\beta_1$&\multicolumn{6}{c}{20}\\
        $\epsilon^t$ &\multicolumn{6}{c}{0.01}\\
        time embedding & \multicolumn{6}{c}{positional}\\
        \# time emb. dims & \multicolumn{6}{c}{16}\\
        SDE type & \multicolumn{6}{c}{variance preserving SDE}\\
        \# layers &3 &5&3&3&3&3\\
        \# hidden dims &64&64&32&32&16&16\\
        \# attention heads &4&4&2&4&2&4\\
        \hline
        \noindent\textbf{Training} & \\
        optimizer & \multicolumn{6}{c}{Adam}\\
        learning rate (VAE) & 1e-4 & 1e-4 & 1e-4 & 1e-4 & 1e-3 & 1e-3\\
        weight decay (VAE) &\multicolumn{6}{c}{1e-4}\\
        learning rate (SGM) & 1e-4 & 1e-4 & 1e-4 & 1e-4 & 1e-3 & 1e-3\\
        weight decay (SGM) &\multicolumn{6}{c}{1e-4}\\
        KL annealing to &0.7&0.7&1.0&1.0&1.0&1.0\\
        \# epochs &1000& 2000& 15000& 15000& 4000& 4000\\
        batch size & 256 & 256 & 4 & 4 & 8 & 8\\
        total training time (h)&72&120&72&72&8&8\\
        \hline
        \noindent\textbf{Evaluation} & \\
        \# used samples &10000&10000&128&128&128&128\\
        ODE solver error tolerance &1e-4&1e-4&1e-4&1e-4&1e-5&1e-5\\
        \hline
    \end{tabular}
    \label{tab:hyperparams}
\end{table*}
\subsection{Cross Entropy}
\label{app:ce}
Directly training the KL term in Eq. (\ref{eq:general-elbo}) is impossible as it involves the marginal score $\nabla_{\bZ^t}\log{q_t(\bZ^t)}$. While the entropy term is fixed to a constant, we show that cross entropy term can be estimated using the following theorem:
 \begin{theorem}{\citep{vahdat2021score}} Given two distributions $q(\bZ^0|\bA,\bX)$ and $p(\bZ^0)$, defined in the continuous space $\bbR^{N\times d}$, denote the marginal distributions of diffused samples under the SDE in Eqn. (\ref{sde}) at time t with $q(\bZ^t|\bA,\bX)$ and $p(\bZ^t)$. Assuming mild smoothness conditions on $\log{q(\bZ^t|\bA,\bX)}$ and $\log{p(\bZ^t)}$, the cross entropy is:
 \begin{align}
     \mathbb{CE}&(q(\bZ^0|\bA,\bX)||p(\bZ^0))\nonumber\\
     &=\bbE_{t,\bepsilon,q_\phi(\bZ^0,\bZ^t|\bA,\bX)}\Big[\frac{g(t)^2}{2}||\bepsilon - \bepsilon_\theta(\bZ^t,t)||^2_2\Big]+\frac{C}{2}, 
 \end{align}
where $t\sim U[0,1]$, $\bepsilon\sim\mathcal{N}(\bepsilon;\bzero,\bI)$, $C=Nd\log{2\pi e\sigma_0^2}$, and $q_\phi(\bZ^0,\bZ^t|\bA,\bX)=q(\bZ^t|\bZ^0)q_\phi(\bZ^0|\bA,\bX)$.

\end{theorem}
Here we focus on variance preserving SDE~(VPSDE)~ \citep{ho2020denoising,song2020score}, and denote $q(\bZ^t|\bZ^0)=\calN(\bZ^t;\bmu_t(\bZ^0),\sigma_t\bI)$, with $\sigma_0=0$ at time step $t=0$. Specifically, the Gaussian parameters in the transition kernel is computed as follow:
\begin{align}
    &\bmu_t(\bZ^0)=e^{\frac{1}{2}\int_0^t\beta(s)\mathrm{d}s}\bZ^0,\nonumber\\
    &\sigma_t = 1-(1-\sigma_0^2)e^{\frac{1}{2}\int_0^t\beta(s)\mathrm{d}s},\nonumber\\
    &\beta(t) = \beta_0+(\beta_1-\beta_0)t,
\end{align}
where $\beta_0$ and $\beta_1$ are hyperparameters denoting the variances at the start and the end of the diffusion process, respectively. For the drift and diffusion coefficients in Eqn. (\ref{sde}), we have $f(t) = -\beta(t)/2$ and $g(t)=\sqrt{\beta(t)}$.

Moreover, as we use VPSDE and have $\sigma_0=0$ at time step $t=0$, we are not able to compute the integral of the cross entropy over the full time interval $[0,1]$. The constant term $Nd/2\log{2\pi e\sigma_0^2}$ Following~\citet{vahdat2021score}, we approximate the cross entropy term by limiting $t\sim U[\epsilon^t,1]$, where $\epsilon^t$ is another hyperparameter, and yield:
\begin{align}
    \mathbb{CE}&(q(\bZ^0|\bA,\bX)||p(\bZ^0))\nonumber\\
    &\approx \bbE_{t,\bepsilon,q_\phi(\bZ^0,\bZ^t|\bA,\bX)}\Big[\frac{g(t)^2}{2}||\bepsilon - \bepsilon_\theta(\bZ^t,t)||^2_2\Big]+\frac{C_{\epsilon^t}}{2},
\end{align}
where $t\sim U[\epsilon^t,1]$, $\bepsilon\sim\mathcal{N}(\bepsilon;\bzero,\bI)$, $C_{\epsilon^t}=Nd\log{2\pi e\sigma_{\epsilon^t}^2}$, and $q_\phi(\bZ^0,\bZ^t|\bA,\bX)=q(\bZ^t|\bZ^0)q_\phi(\bZ^0|\bA,\bX)$. Since the SGM prior never learns to recover $\bZ^0$ exactly, the stage for finetuning the decoder by injecting extra noise becomes necessary when the sample quality heavily relies on the reconstruction accuracy. Empirically, we find that the finetune stage improves the validity metric by 15\% on the ZINC250K dataset.

\subsection{Exchangeability of NVDiff}
We guarantee the encoder and the decoder are permutation-equivariant by utilizing GNNs. And since the SGM prior is designed to be an exchangeable distribution, 
NVDiff is guaranteed to be an exchangeable model.

\begin{table*}[h!]
    \centering
    \caption{Dataset statistics}
    \begin{tabular}{lcccccc}
    \hline
    &\multicolumn{2}{c}{Molecular dataset} & \multicolumn{4}{c}{Generic graph dataset}\\
         &  QM9&ZINC250K&Community&Ego&Community-small&Ego-small\\\hline
        \# graphs & 130828&249456&500&753&500&500\\
        \# nodes(min,max) & (2, 9)&(6, 38)&(60, 160)&(50, 399)&(12, 20)&(4, 18)\\
        \# edges(min,max) & (1, 14)&(5, 45)&(240, 1995)&(57, 1071)&(23, 82)&(3, 63)\\
        \# node types &4&9&1&1&1&1\\ 
        \# edge types &4&4&1&1&1&1\\\hline
    \end{tabular}
    \label{tab:data-stats}
\end{table*}

\begin{table*}[h]
    \centering
    \caption{Further experiment results on molecule generation. Results of baselines marked with ``*'' are taken from previous papers. We used paired t-test to compare the results; the numbers in bold indicate that the method is better at the 5\% significance level; the number underlined are the second best.}
    \begin{tabular}{lcccc}\hline
    \multicolumn{5}{c}{QM9}\\
    &Validity$\uparrow$& Uniqueness$\uparrow$& NSPDK$\downarrow$ & FCD$\downarrow$ \\\hline
    GraphDF$^*$ & 82.67 & 97.62 &  0.063$\pm$0.001& 10.816$\pm$0.020\\
    Moflow$^*$ & 91.36$\pm$1.23&{94.72$\pm$0.77}&0.017$\pm$0.003&4.467$\pm$0.595\\
    GraphCNF & \underline{95.00$\pm$1.15}&93.68$\pm$0.96&\textbf{0.002$\pm$0.000}&1.629$\pm$0.326\\
    EDP-GNN$^*$ & 47.52$\pm$3.60&\textbf{99.25$\pm$0.05}&0.005$\pm$0.001&2.680$\pm$0.221\\
    GDSS$^*$ & \underline{95.72$\pm$1.94}&\underline{98.46$\pm$0.61}&\underline{0.003$\pm$0.000}&2.900$\pm$0.282\\\hline
    NVDiff & \underline{95.79$\pm$1.32}&96.53$\pm$1.16&\textbf{0.002$\pm$0.000}&\underline{1.131$\pm$0.213}\\
    NVDiff(10\%) &90.29$\pm$1.12&94.88$\pm$0.96& \underline{0.003$\pm$0.000} &1.758$\pm$0.390\\
    NVDiff-E & \textbf{98.49$\pm$0.81}&\underline{98.32$\pm$0.55}&\textbf{0.002$\pm$0.000}&\textbf{0.350$\pm$0.122}\\\hline
    \multicolumn{5}{c}{ZINC250K}\\
    &Validity$\uparrow$ & Uniqueness$\uparrow$ & NSPDK$\downarrow$ & FCD$\downarrow$\\\hline
    GraphDF$^*$ &  \underline{89.03}& 99.16&  0.176$\pm$0.001&34.202$\pm$0.160 \\
    Moflow$^*$ &63.11$\pm$5.17 &  \textbf{100$\pm$0.00} &\underline{0.046$\pm$0.002}& 20.931$\pm$0.184  \\
    GraphCNF &\textbf{95.26$\pm$1.85} &\underline{99.96$\pm$0.03}&0.089$\pm$0.002&14.786$\pm$1.326\\
    EDP-GNN$^*$ & 82.97$\pm$2.73 & 99.79$\pm$0.08 & \underline{0.049$\pm$0.006} &16.737$\pm$1.300\\
    GDSS$^*$ & \textbf{97.01$\pm$0.77} & 99.64$\pm$0.13 &  \textbf{0.019$\pm$0.001} & 14.656$\pm$0.680\\\hline
    NVDiff & 85.63$\pm$0.89 &\textbf{99.99$\pm$0.01} & 0.066$\pm$0.006 & \textbf{4.019$\pm$0.01}\\
    NVDiff-E&\textbf{96.23$\pm$1.21}&\underline{99.94$\pm$0.03}&0.061$\pm$0.001&\underline{6.681$\pm$0.01}\\\hline
    \end{tabular}
    \label{tab:add-chem}
\end{table*}
\begin{table*}[h]
\begin{minipage}[b]{0.45\textwidth}   
\caption{Ablation study of contextual vector on QM9 and ZINC250K.}
        {\begin{tabular}{ccccc}\hline
             & \multicolumn{2}{c}{QM9} &\multicolumn{2}{c}{ZINC250K}\\
             & Validity & FCD& Validity&FCD\\\hline
            w/o $g_0$ &93.18 & 3.082& 83.11& 6.100\\
            w/ $g_0$ & 95.79 & 1.131 & 85.63& 4.019\\\hline
        \end{tabular}}
    \label{tab:context-abl}
\end{minipage}
\hspace{2em}
\begin{minipage}[b]{0.45\textwidth}  
\caption{Ablation study of consistence regularization on QM9 and ZINC250K.}
        {\begin{tabular}{ccccc}\hline
             & \multicolumn{2}{c}{QM9} &\multicolumn{2}{c}{ZINC250K}\\
             & Validity & FCD& Validity&FCD\\\hline
            w/o $g_0$ &94.24 & 1.021& 71.35& 4.295\\
            w/ $g_0$ & 95.79 & 1.131 & 85.63& 4.019\\\hline
        \end{tabular}}
    \label{fig:consist-abl}
\end{minipage}
\end{table*}

\section{Dataset Statistics}
\label{app:data-stats}
Table \ref{tab:data-stats} shows the details of the used datasets. We use 90\% of each dataset for training and test the rest for molecular datasets. For generic graph datasets, we use 80\% of the dataset for training. 
\section{Experimental Details}
\label{app:exp-detail}
All hyperparameters of NVDiff on different datasets are provided in Table \ref{tab:hyperparams}.
\subsection{Training}
We mostly follow the previous works to optimize NVDiff. We pretrain the VAE with a fixed prior before warming up the encoder and decoder before replacing the SGM prior. During the pretraining, KL annealing is also applied. We adopt the importance sampling scheme proposed by~\citep{vahdat2021score} to sample time step t to reduce the gradient variance, thus enabling faster convergence. Following~\citet{sinha2021d2c}, we normalize the latent node vectors obtained from the encoder before feeding it into the SGM prior. We did not observe improvement by utilizing an exponential moving average (EMA) for SGM prior and VAE parameters during test time. We also apply gradient clipping during the optimization. For the edge distribution in the decoder, we optimize the edge-existence distribution $p(\ba_{i,j}|\br_{i,j})$ and the edge-type distribution $p(\ba_{i,j}|\ba_{i,j}\neq\bzero,\br_{i,j})$ at the same time. Specifically, we fit the edge-existence distribution with the binary adjacency matrix and the edge-type distribution for all edge entries weighted by the ground-truth edge-existence indicator. During finetuning, we fix the parameters of the encoder and only train the decoder for more epochs by optimizing $\calL_\mathrm{Reg}$. 

\subsection{Evaluation}
We report all results over 5 runs with different random seeds for each dataset. 

\noindent\textbf{Molecule generation.} We precompute the prior distribution of graph size over the training set. The graph size distribution is then used to draw the number of nodes each sample should contain. We report Uniqueness, NSPDK MMD, and FCD metrics over the valid samples. We do not employ any validity correction scheme during training or sampling ~\citep{shi2020graphaf, luo2021graphdf}.

\noindent\textbf{Generic graph generation.} We compute MMD between the 128 generated graphs and graphs in the test set. We mainly follow the configurations in \citet{you2018graphrnn, thompson2022evaluation} when evaluating MMD. We extract the largest sub-component of the generated graphs when evaluating Ego and Ego-small. 

\subsection{Configuration of Downstream tasks for Contextual Vector}
We use QM9 for contextual vector analysis. Here we specify the setups for each downstream tasks. We record the contextual vectors over all training graphs for t = [0, 0.05, 0.1, 0.15, 0.2, 0.25, 0.3, 0.35, 0.4, 0.45, 0.5, 0.55, 0.6, 0.65, 0.7, 0.75, 0.8, 0.85, 0.9, 0.95, 1.0]. For each intermediate time step t, We train and evaluate on the collected contextual vectors. We use 90\% of the data for training and 10\% of the data for testing. For all tasks, we use a 3-layer MLP.

\noindent\textbf{Counting atom types.} QM9 contains four heavy atoms (O, N, C, F). Here we manually generate the label of the number of atoms of the four types for each molecule. We consider the counting problem as each atom type as an independent task. We report the MAE metric by averaging over four counting tasks.

\noindent\textbf{Cycle detection and diameter prediction.}
We only consider a binary classification task for cycle detection and use an MLP to determine whether a molecule contains a cycle. We ignore the atom type and bond type when computing the existence of the cycle and the diameter of the molecules. We perform a regression task on the diameter prediction.
\subsection{Computational Resources}
We train and evaluate NVDiff on GeForce RTX 3060, A100, or RTX 6000 GPU.

\section{Additional Results}
\label{app:extra-exp}
\subsection{Additional Results on Generic Graph Generation}
We additionally provide the standard derivation of the all the results for generic graph generation in Table \ref{tab:generic-std}.
\subsection{Additional Metrics on Molecule Generation}
We additionally provide the Uniqueness, and NSPDK metrics as well as all standard derivations of the results in Table \ref{tab:add-chem}. We also report the performance of the NVDiff-E. 


\begin{table*}[h]
    \centering
    \caption{ Further experiment results on generic graph generation. Results of baselines marked with ``*'' are taken from previous papers. Numbers in bold indicate that the methods are the best, numbers underlined are the second best. }
    
\begin{tabular}{lcccc}
\hline
              \multicolumn{5}{c}{Community-small}                                                                                                                                                               \\
             & Deg.$\downarrow$                          & Clus.$\downarrow$                         & Orbit$\downarrow$                         & {NN$\downarrow$}                            \\ \hline
GraphRNN$^*$ & 0.08                                      & 0.120                                     & 0.040                                     & 0.997$\pm$0.010 \\
GRAN         & 0.070$\pm$0.009                           & \underline{0.045$\pm$0.004}                           & 0.021$\pm$0.003                           & \underline{0.701$\pm$0.093}                                                \\
GraphCNF     & \textbf{0.021$\pm$0.001} & 0.141$\pm$0.004                           & 0.044$\pm$0.006                           & 0.876$\pm$0.138                                                \\
EDP-GNN$^*$  & 0.053                                     & 0.144                                     & 0.026                                     & 0.798$\pm$0.159                                                \\
GDSS$^*$     & \underline{0.045$\pm$0.028}                           & 0.086$\pm$0.022                           & \textbf{0.007$\pm$0.004} & 0.858$\pm$0.166                                                \\ \hline
NVDiff       & \textbf{0.021$\pm$0.001} & \textbf{0.035$\pm$0.002} & \underline{0.018$\pm$0.002}                           & \textbf{0.688$\pm$0.144}                                               \\ \hline
              \multicolumn{5}{c}{Community}                                                                                                                                                                     \\
             & Deg.$\downarrow$                          & Clus.$\downarrow$                         & Orbit$\downarrow$                         & {NN$\downarrow$}                            \\ \hline
GraphRNN$^*$ & \textbf{0.014}           & \textbf{0.002}           & \underline{0.039}                                     & 1.452$\pm$0.092                                                \\
GRAN         & \underline{0.085$\pm$0.007}                           & \underline{0.068$\pm$0.008}                           & 0.042$\pm$0.003                           & 1.406$\pm$0.191                                                \\
GraphCNF     & 0.707$\pm$0.129                           & 1.305$\pm$0.326                           & 0.194$\pm$0.062                           & 1.356$\pm$0.261                                                \\
EDP-GNN      & 0.137$\pm$0.042                           & 1.104$\pm$0.036                           & 0.051$\pm$0.019                           & 1.416$\pm$0.133                                                \\
GDSS         & 0.099$\pm$0.009                           & 0.327$\pm$0.023                           & 0.162$\pm$0.057                           & \underline{1.203$\pm$0.353}                                                \\ \hline
NVDiff       & 0.093$\pm$0.008                           & 0.086$\pm$0.006                           & \textbf{0.021$\pm$0.001} & \textbf{0.982$\pm$0.155}                                                \\ \hline
              \multicolumn{5}{c}{Ego-small}                                                                                                                                                                     \\
             & Deg.$\downarrow$                          & Clus.$\downarrow$                         & Orbit$\downarrow$                         & {NN$\downarrow$}                            \\\hline
GraphRNN$^*$ & 0.090                                     & 0.220                                     & \underline{0.003}                                     & \textbf{1.094$\pm$0.003}                                                \\
GRAN         & 0.020$\pm$0.001                           & 0.126$\pm$0.021                           & 0.010$\pm$0.002                           & \textbf{1.095$\pm$0.003}                                                \\
GraphCNF     & \underline{0.011$\pm$0.001}                           & \textbf{0.011$\pm$0.001} & \textbf{0.001$\pm$0.000} & \textbf{1.094$\pm$0.002}                                                \\
EDP-GNN$^*$  & 0.052                                     & 0.093                                     & 0.007                                     & \textbf{1.095$\pm$0.002}                                                \\
GDSS$^*$     & 0.021$\pm$0.008                           & \underline{0.024$\pm$0.007}                           & 0.007$\pm$0.004                           & \underline{1.097$\pm$0.001}                                                \\\hline
NVDiff       & \textbf{0.005$\pm$0.001} & 0.045$\pm$0.004                           & \textbf{0.001$\pm$0.000} & \textbf{1.092$\pm$0.002}                                                \\\hline
              \multicolumn{5}{c}{Ego}                                                                                                                                                                           \\
             
             & Deg.$\downarrow$                          & Clus.$\downarrow$                         & Orbit$\downarrow$                         & {NN$\downarrow$}                            \\\hline
GraphRNN$^*$ & 0.077                                     & 0.316                                     & \underline{0.030}                                     & \underline{0.192$\pm$0.002}                                                \\
GRAN         & \underline{0.064$\pm$0.012}                           & \underline{0.279$\pm$0.046}                           & 0.039$\pm$0.009                           & 0.196$\pm$0.008                                                \\
GraphCNF     & OOM                                       & OOM                                       & OOM                                       & OOM                                                           \\
EDP-GNN$^*$  & 0.069$\pm$0.010                           & 0.447$\pm$0.065                           & 0.055$\pm$0.014                           & \textbf{0.189$\pm$0.001}                                                \\
GDSS$^*$     & 0.134$\pm$0.012                           & 0.371$\pm$0.042                           & 0.148$\pm$0.02                            & 0.295$\pm$0.069                                                \\\hline
NVDiff       & \textbf{0.045$\pm$0.004} & \textbf{0.091$\pm$0.011} & \textbf{0.004$\pm$0.001} & 0.195$\pm$0.004     \\\hline                                          
\end{tabular}
    \label{tab:generic-std}
\end{table*}
\subsection{{Ablation Studies}}
{Here, we present two ablation studies investigating the effectiveness of contextual embedding and consistency regularization term, respectively. We examine the two components using molecular datasets.}
\paragraph{{Contextual vector.}} {We train NVDiff with diffusion prior without the contextual embedding $g_0$ on two molecular datasets and report the Validity and FCD scores. From Table \ref{tab:context-abl}, we can observe that using contextual embedding improves the validity and FCD metrics.}
\paragraph{{Consistency regularization.}}   {The consistency regularization term improves the robustness of the decoder given noisy samples generated by the diffusion prior. Thus, regularization helps improve the reconstruction precision, which is especially valuable in drug synthesis. We report the effect of the consistency regularization term on the validity metrics over the two molecular datasets. As we can see in Table \ref{fig:consist-abl}, by introducing the consistency regularization term, the model is able to generate molecules with higher validity.}

\subsection{Visualization of Generated Molecules}
\begin{figure*}[h]
    \centering
    \begin{tabular}{ccccc}
        Reference &NVDiff & GraphDF & Moflow &  GraphCNF\\
        {\includegraphics[width=0.10\textwidth]{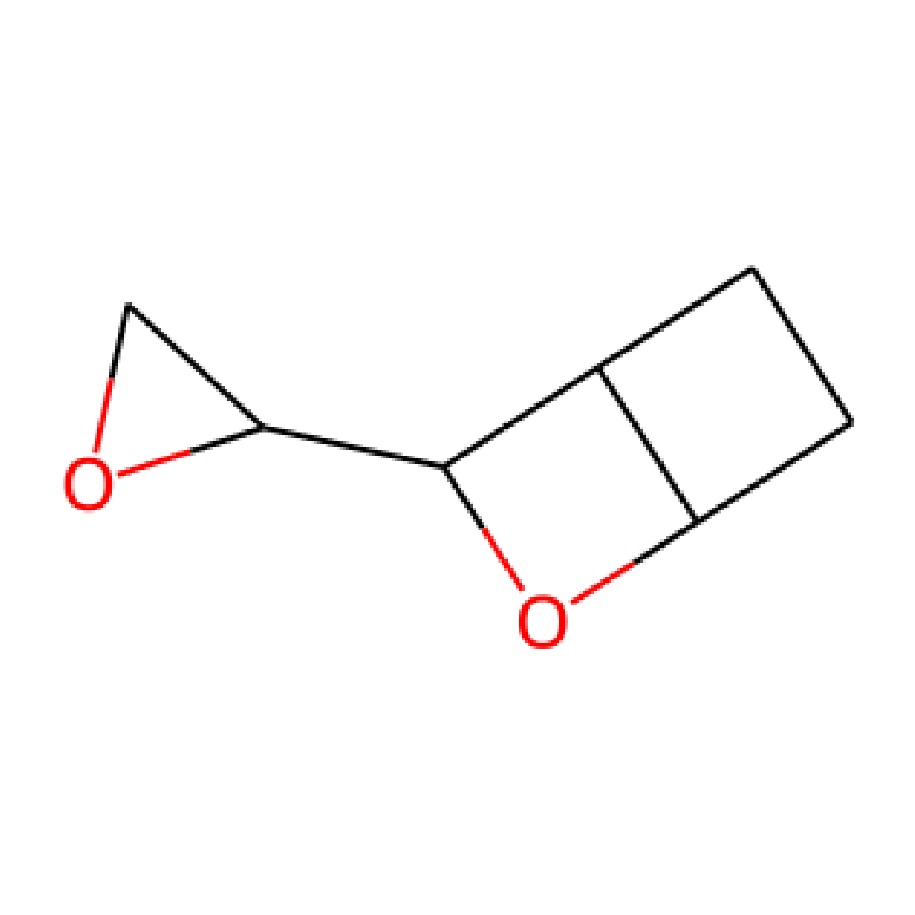}}& 
        {\includegraphics[width=0.185\textwidth]{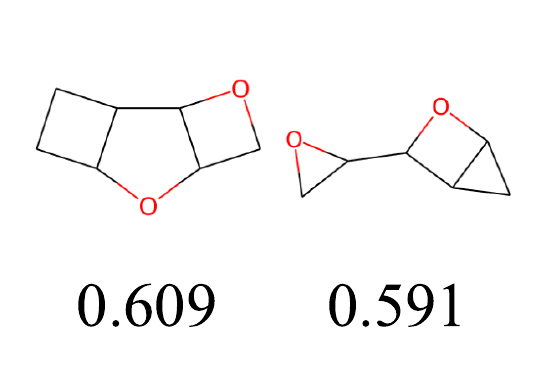}}&
        {\includegraphics[width=0.185\textwidth]{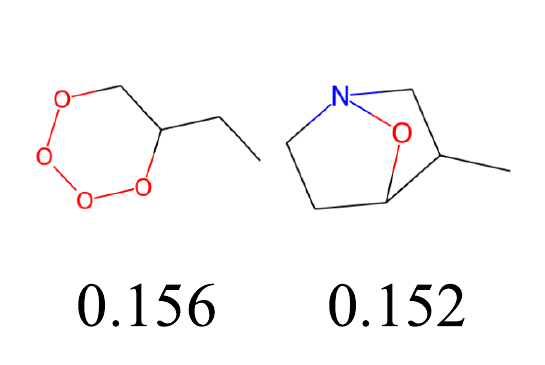}}&
        {\includegraphics[width=0.185\textwidth]{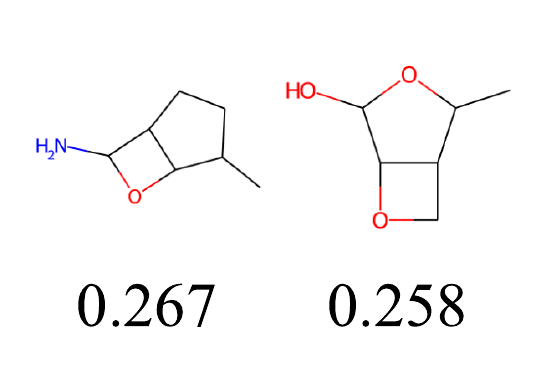}}&
        {\includegraphics[width=0.185\textwidth]{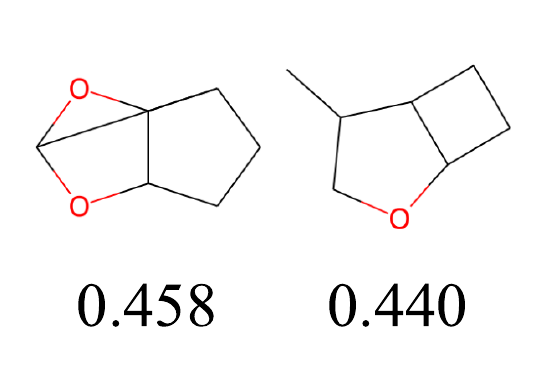}}\\
        {\includegraphics[width=0.10\textwidth]{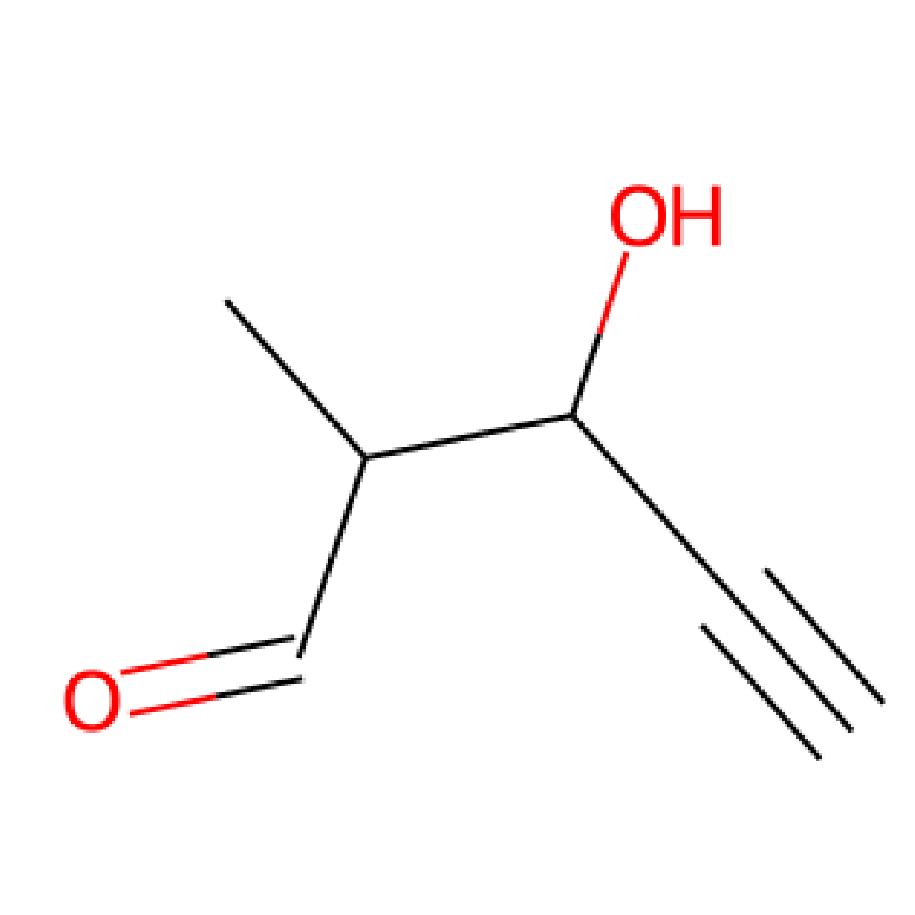}}& 
        {\includegraphics[width=0.185\textwidth]{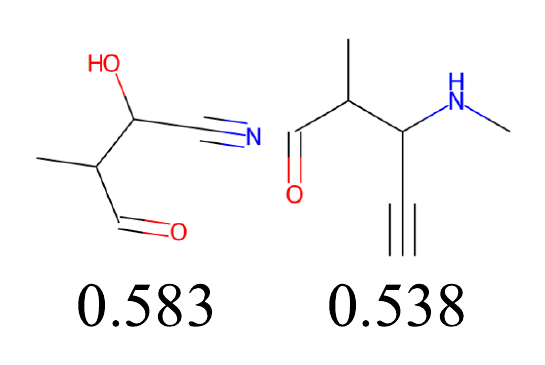}}&
        {\includegraphics[width=0.185\textwidth]{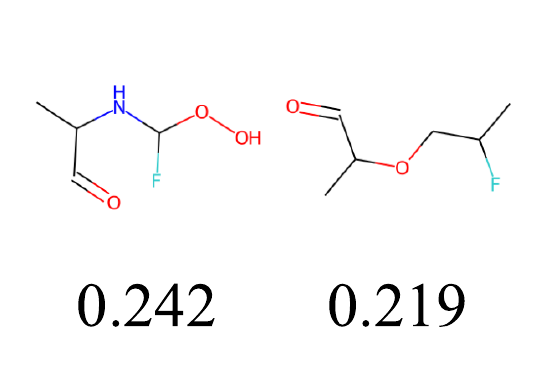}}&
        {\includegraphics[width=0.185\textwidth]{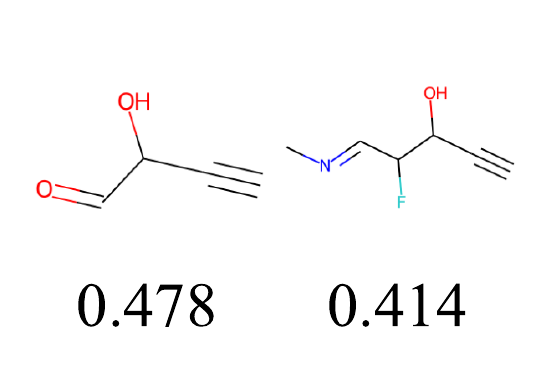}}&
        {\includegraphics[width=0.185\textwidth]{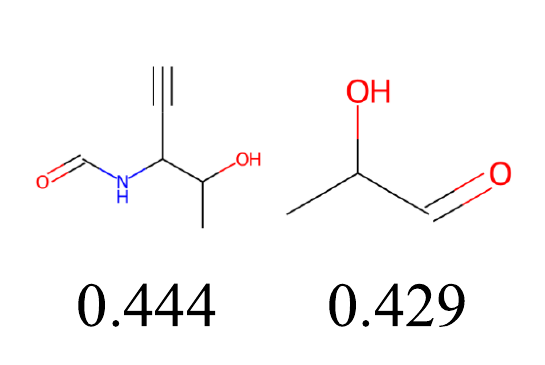}}\\
        {\includegraphics[width=0.10\textwidth]{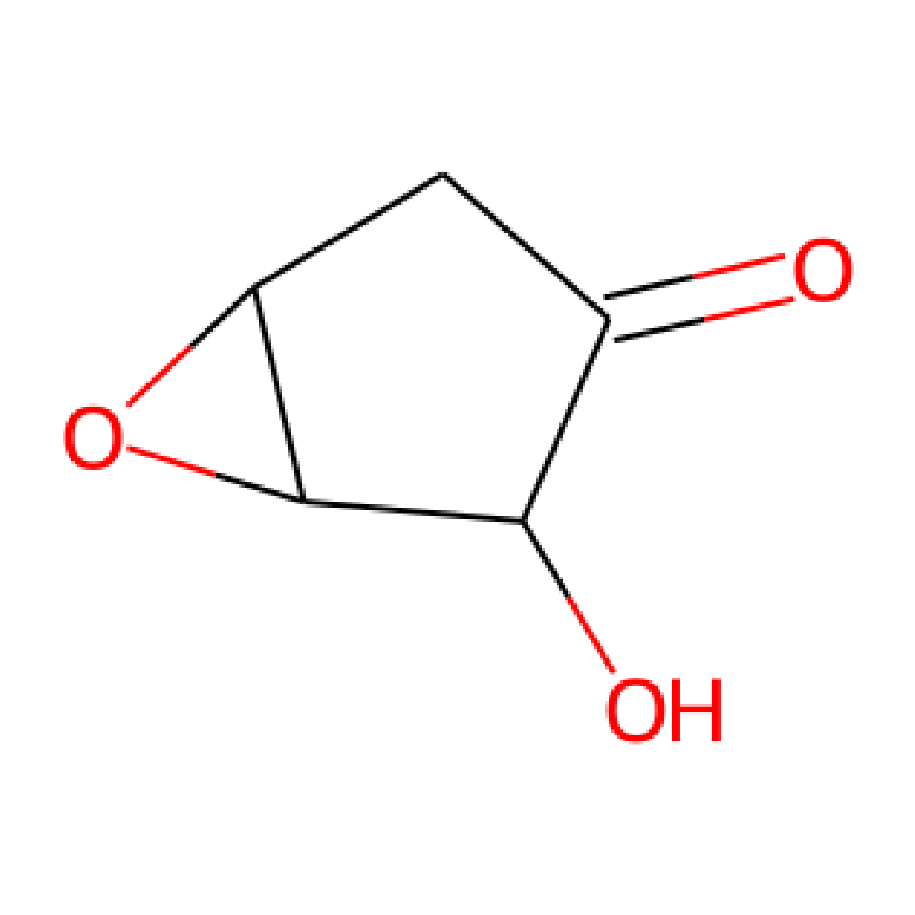}}& 
        {\includegraphics[width=0.185\textwidth]{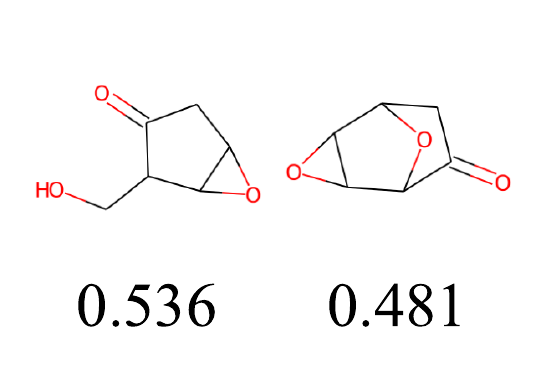}}&
        {\includegraphics[width=0.185\textwidth]{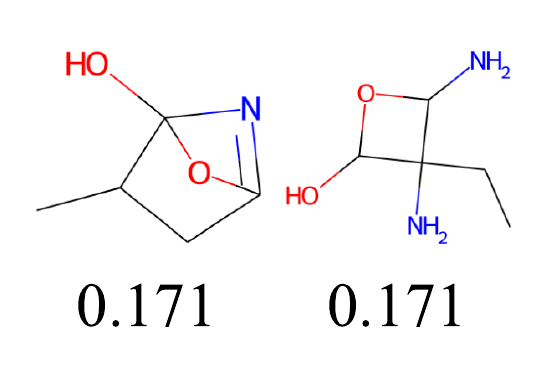}}&
        {\includegraphics[width=0.185\textwidth]{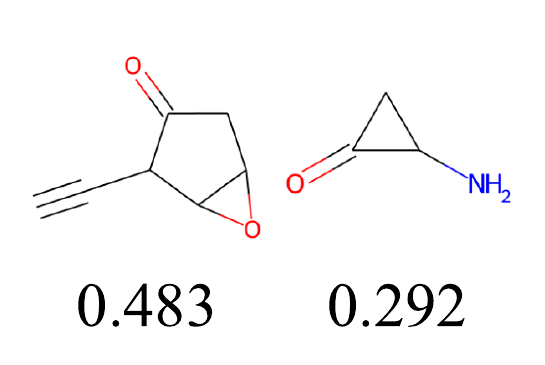}}&
        {\includegraphics[width=0.185\textwidth]{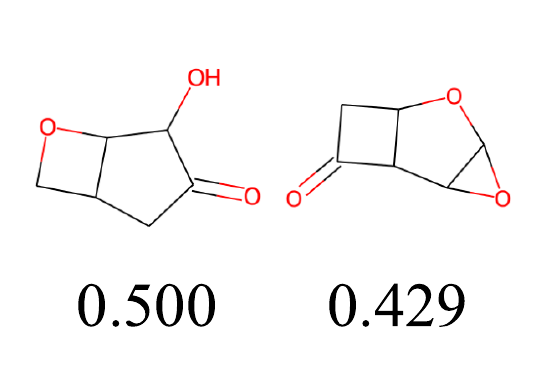}}\\
        {\includegraphics[width=0.10\textwidth]{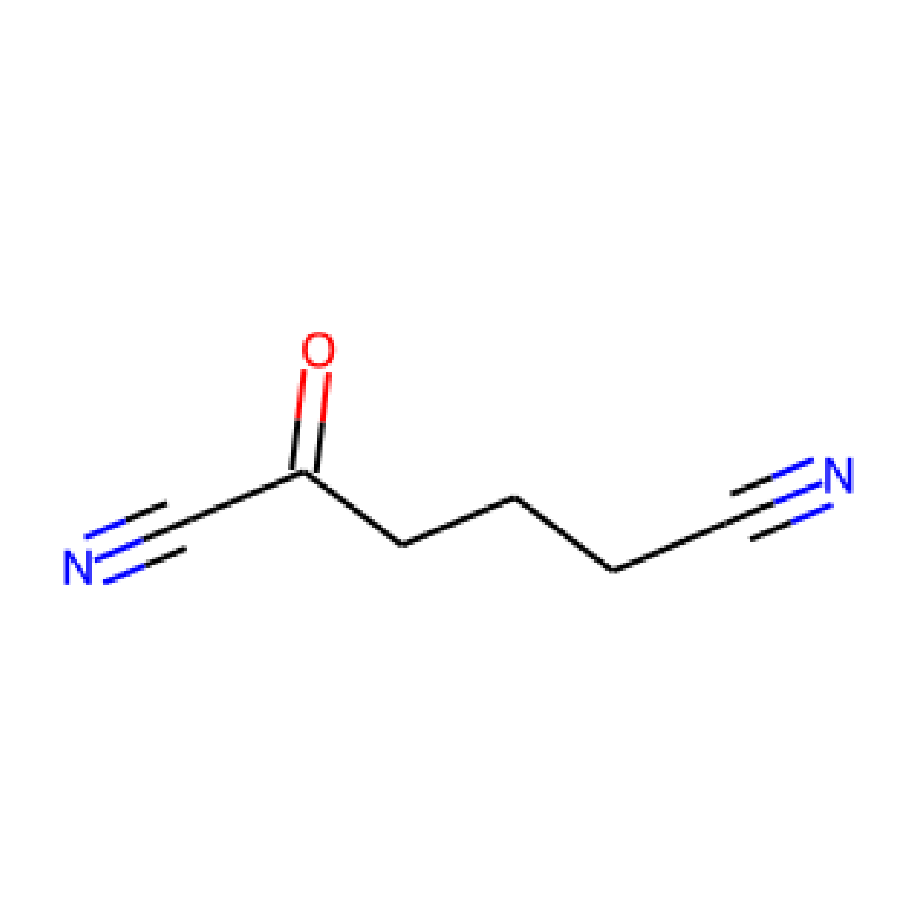}}& 
        {\includegraphics[width=0.185\textwidth]{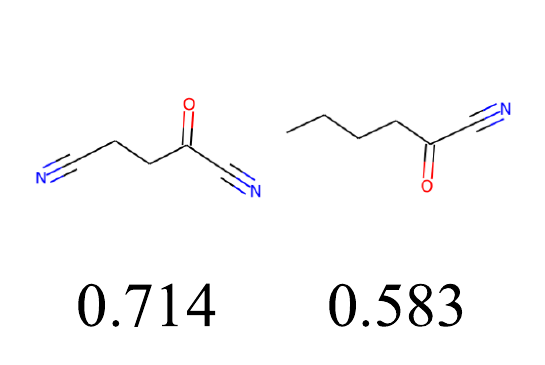}}&
        {\includegraphics[width=0.185\textwidth]{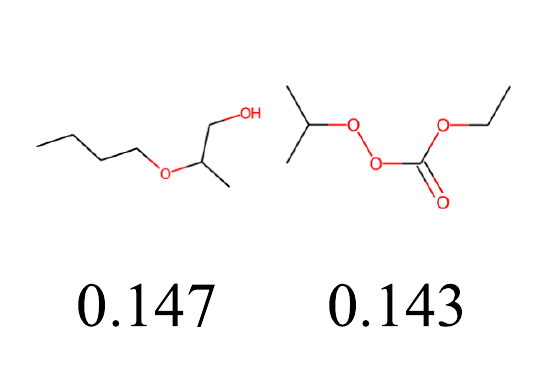}}&
        {\includegraphics[width=0.185\textwidth]{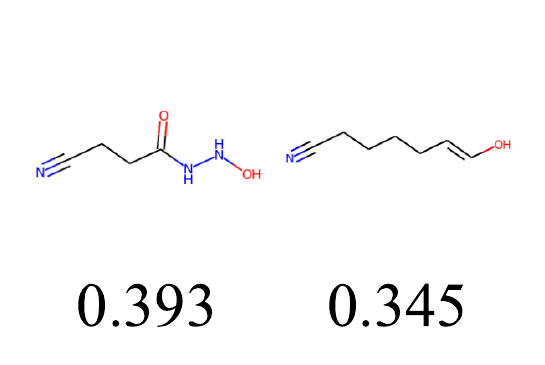}}&
        {\includegraphics[width=0.185\textwidth]{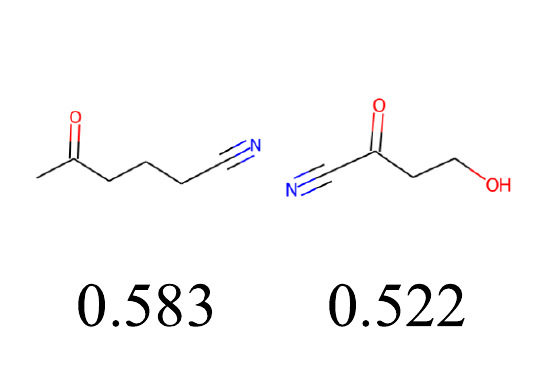}}\\
        {\includegraphics[width=0.10\textwidth]{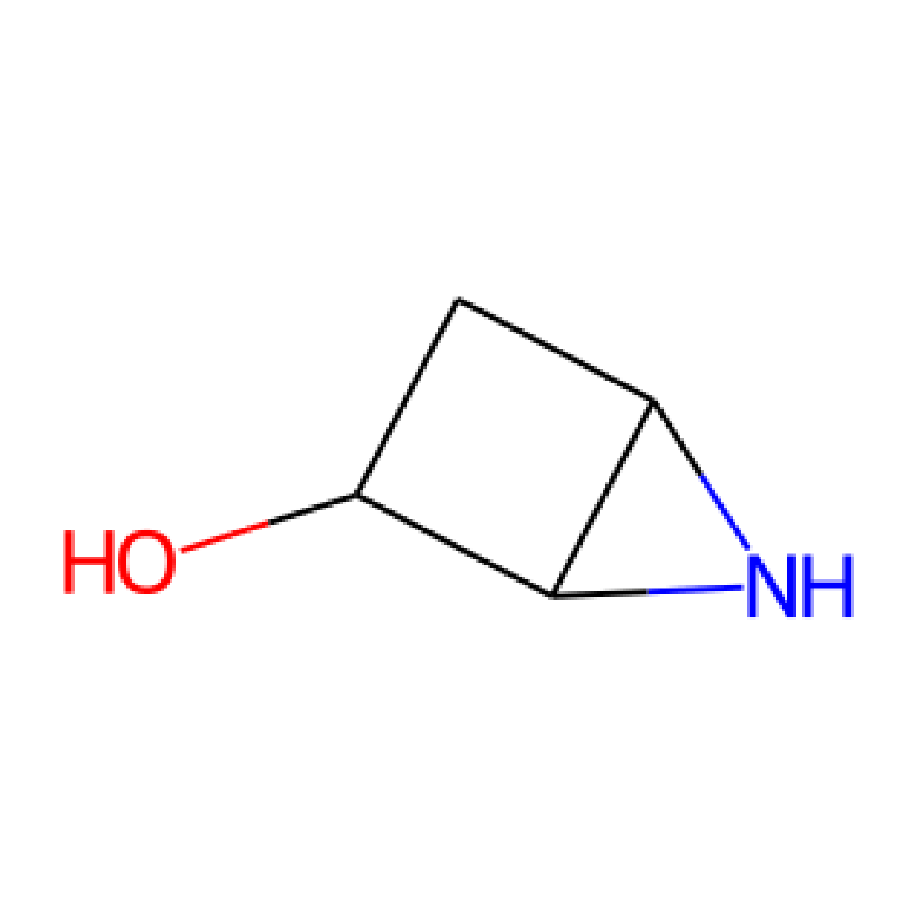}}& 
        {\includegraphics[width=0.185\textwidth]{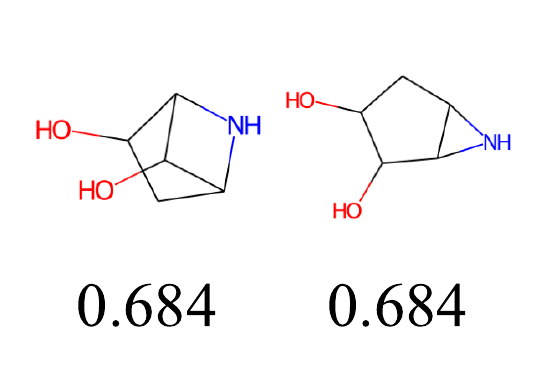}}&
        {\includegraphics[width=0.185\textwidth]{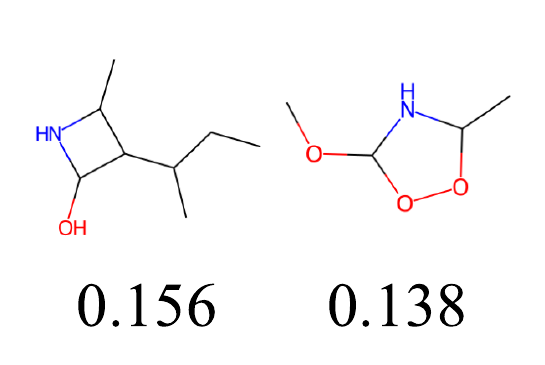}}&
        {\includegraphics[width=0.185\textwidth]{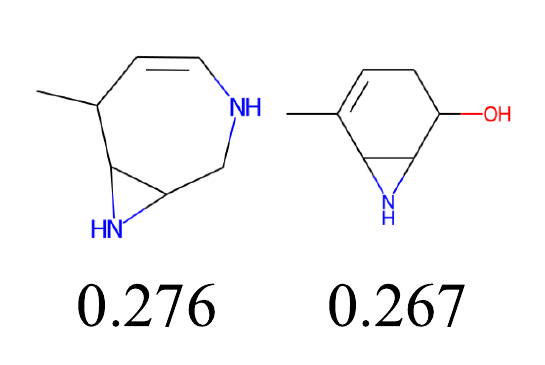}}&
        {\includegraphics[width=0.185\textwidth]{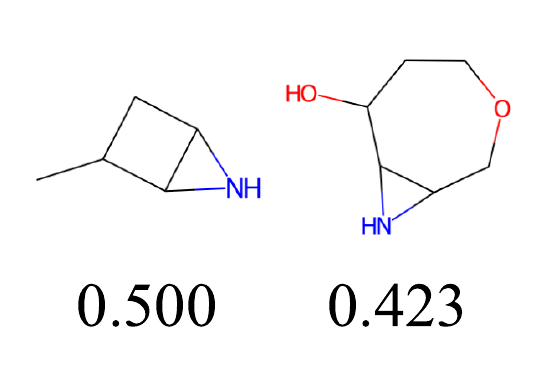}}\\
        {\includegraphics[width=0.10\textwidth]{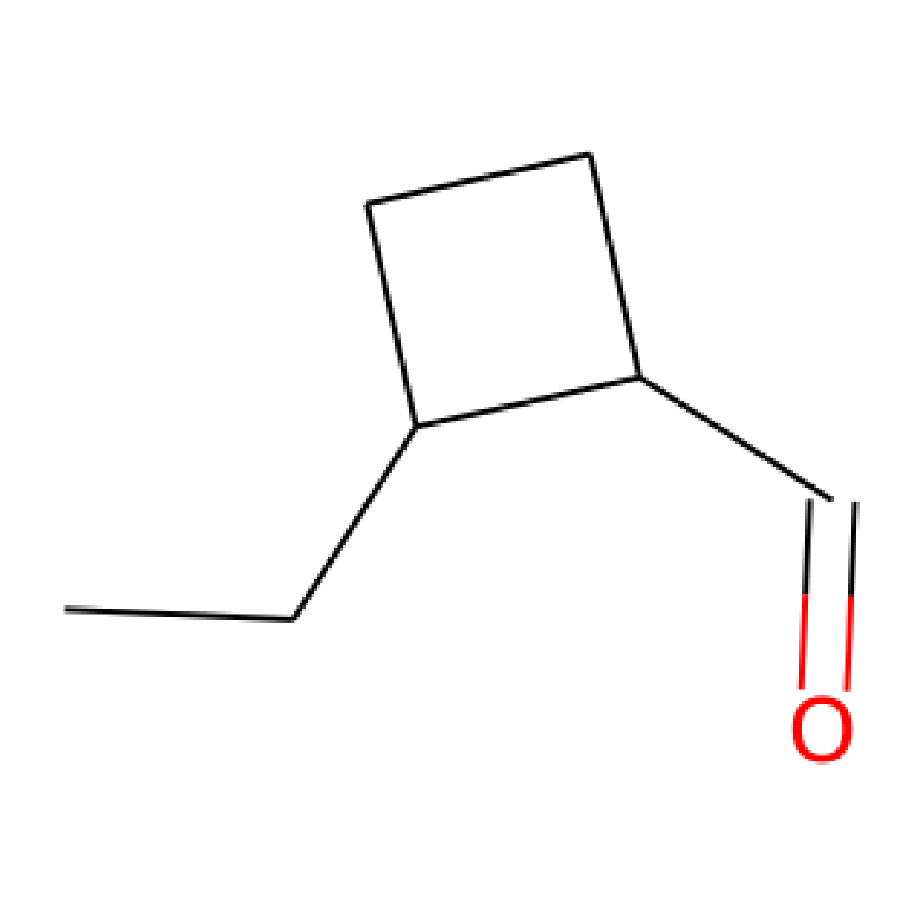}}& 
        {\includegraphics[width=0.185\textwidth]{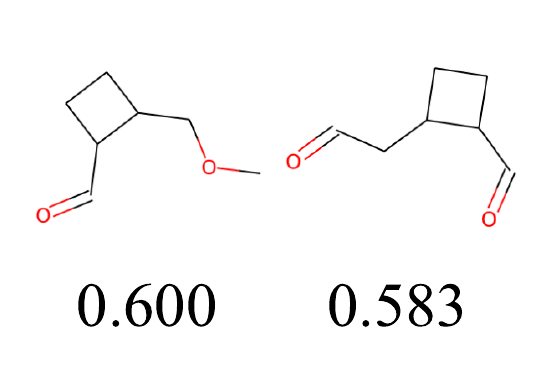}}&
        {\includegraphics[width=0.185\textwidth]{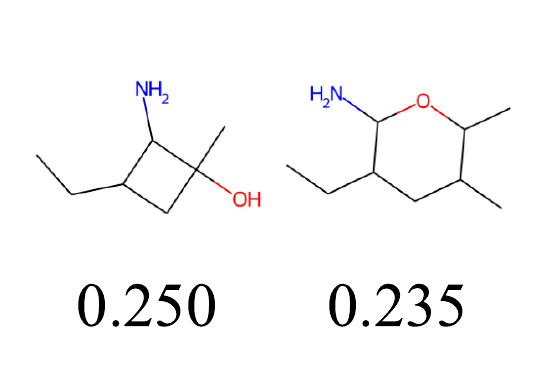}}&
        {\includegraphics[width=0.185\textwidth]{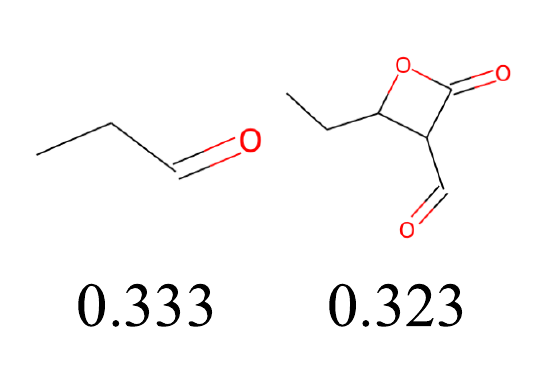}}&
        {\includegraphics[width=0.185\textwidth]{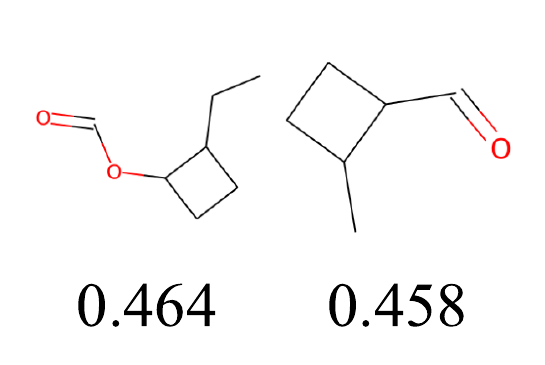}}\\
        {\includegraphics[width=0.10\textwidth]{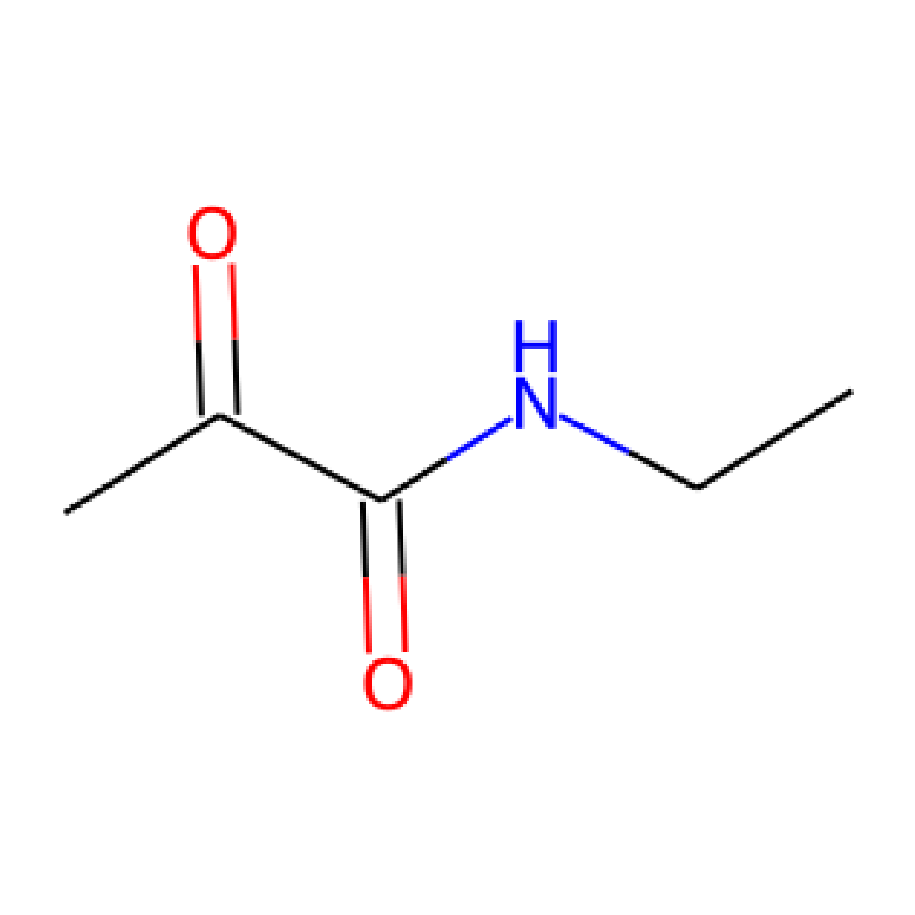}}& 
        {\includegraphics[width=0.185\textwidth]{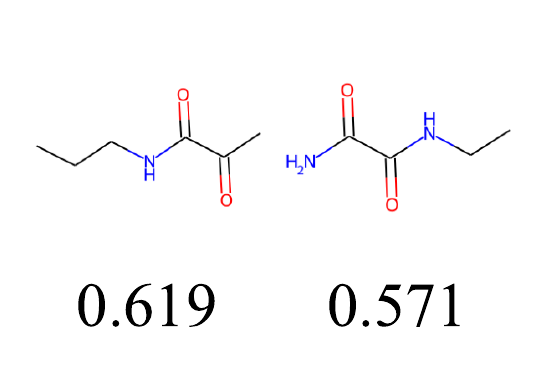}}&
        {\includegraphics[width=0.185\textwidth]{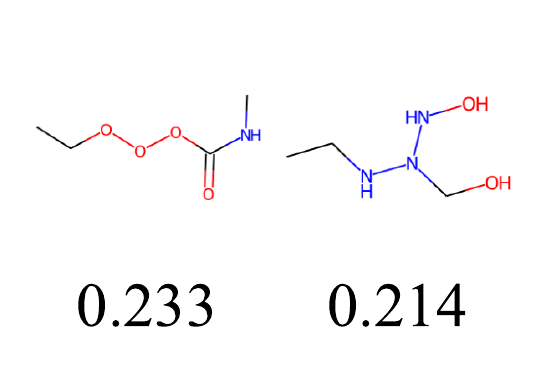}}&
        {\includegraphics[width=0.185\textwidth]{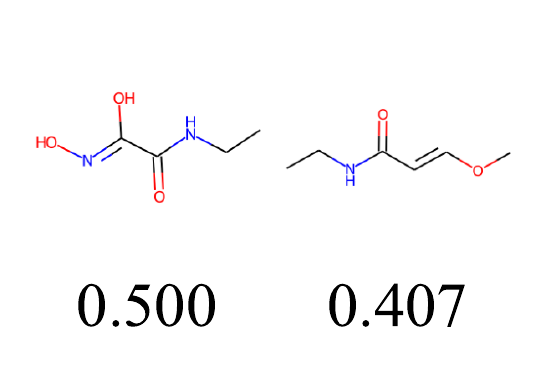}}&
        {\includegraphics[width=0.185\textwidth]{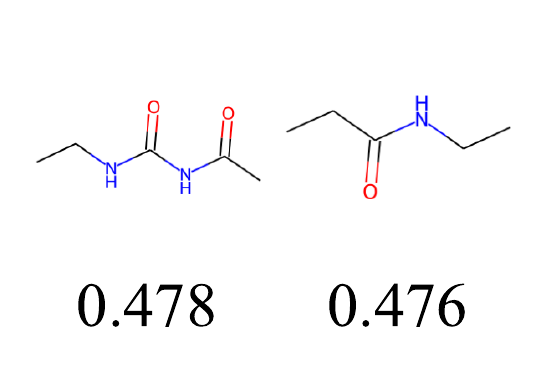}}\\
        {\includegraphics[width=0.10\textwidth]{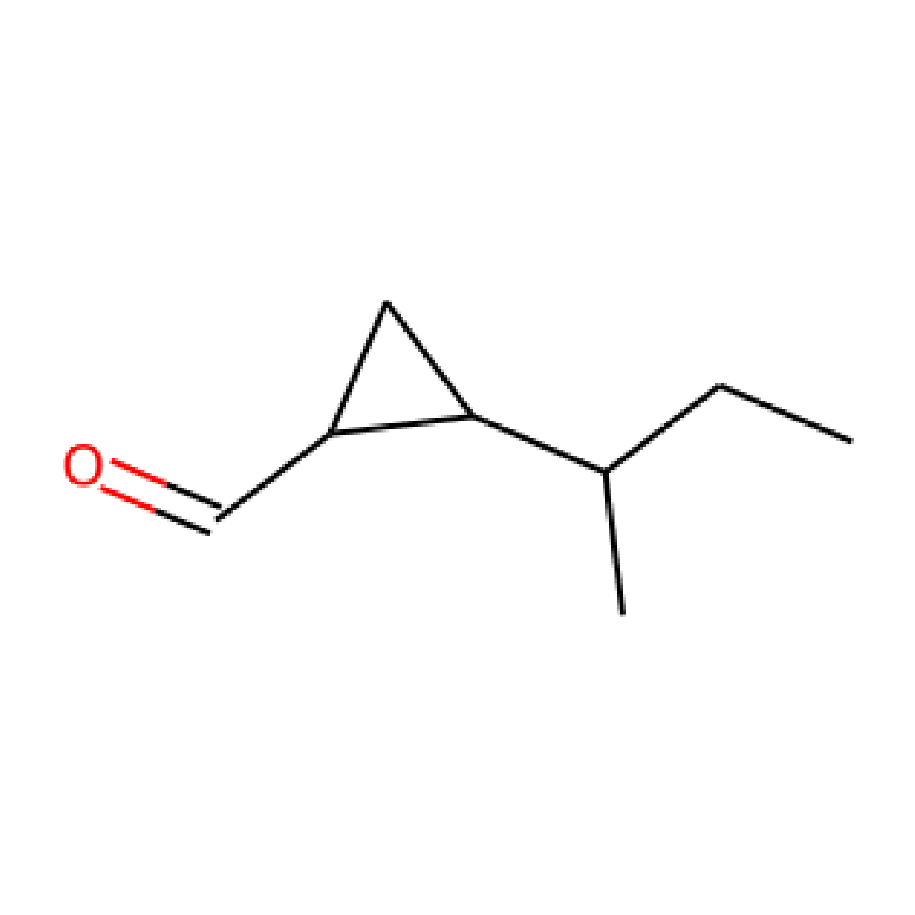}}& 
        {\includegraphics[width=0.185\textwidth]{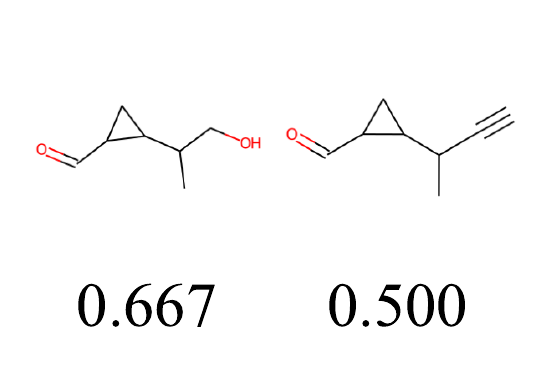}}&
        {\includegraphics[width=0.185\textwidth]{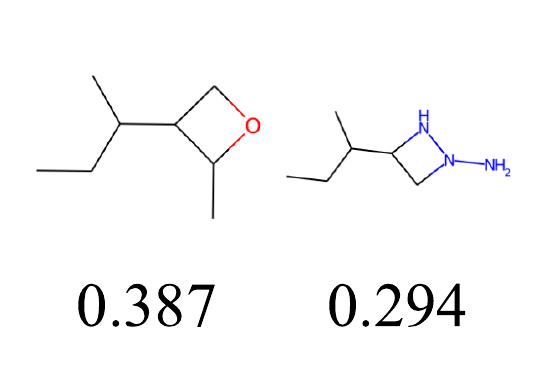}}&
        {\includegraphics[width=0.185\textwidth]{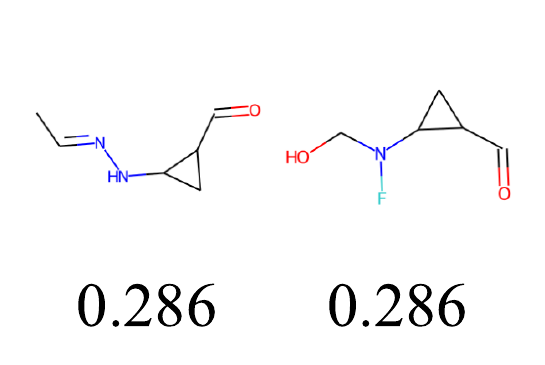}}&
        {\includegraphics[width=0.185\textwidth]{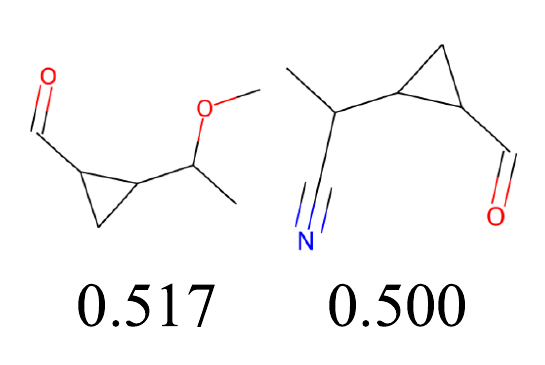}}\\
      \end{tabular}
      
    \caption{Extended result of molecule visualization on QM9.}
    \label{fig:mol_vis_qm9}
\end{figure*}

\begin{figure*}[h]
    \centering
    \begin{tabular}{ccccc}
        Reference &NVDiff & GraphDF & Moflow &  GraphCNF\\
        {\includegraphics[width=0.10\textwidth]{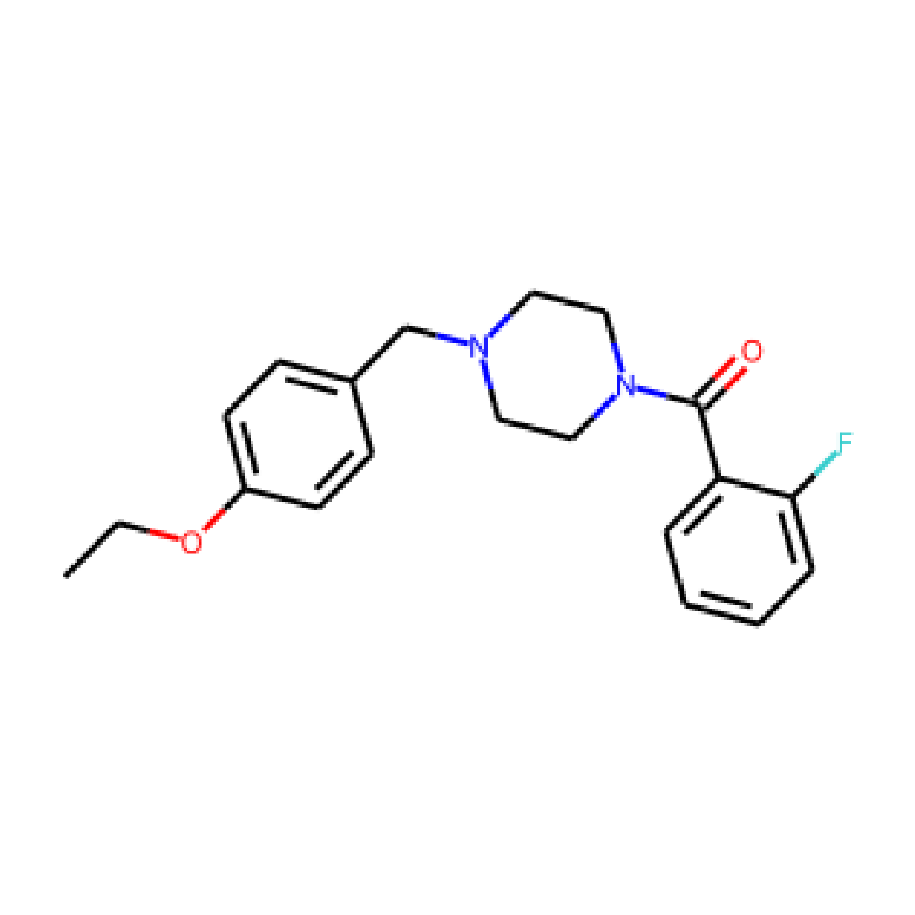}}& 
        {\includegraphics[width=0.185\textwidth]{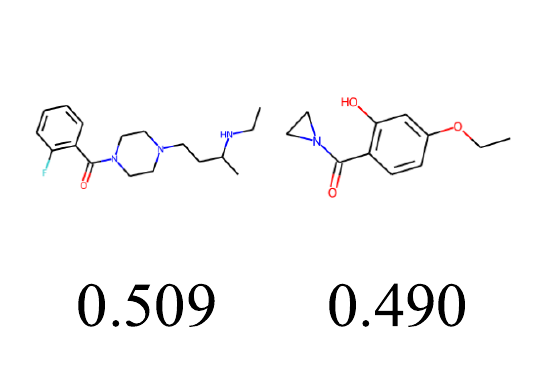}}&
        {\includegraphics[width=0.185\textwidth]{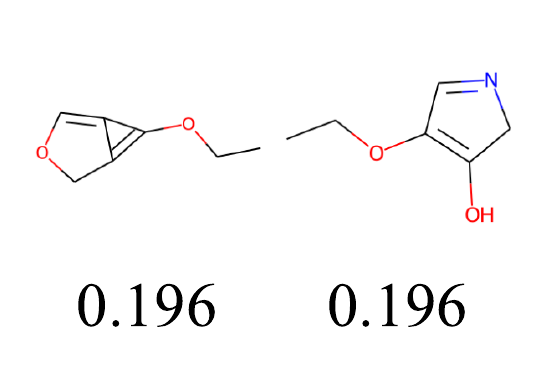}}&
        {\includegraphics[width=0.185\textwidth]{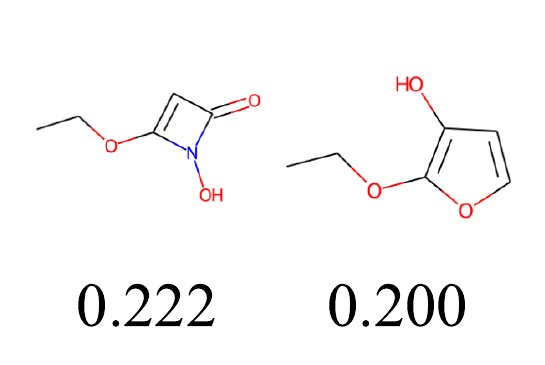}}&
        {\includegraphics[width=0.185\textwidth]{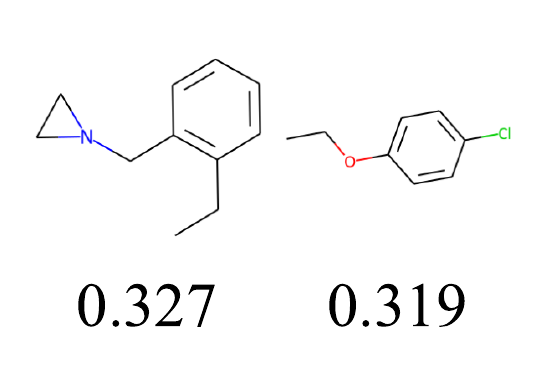}}\\
        {\includegraphics[width=0.10\textwidth]{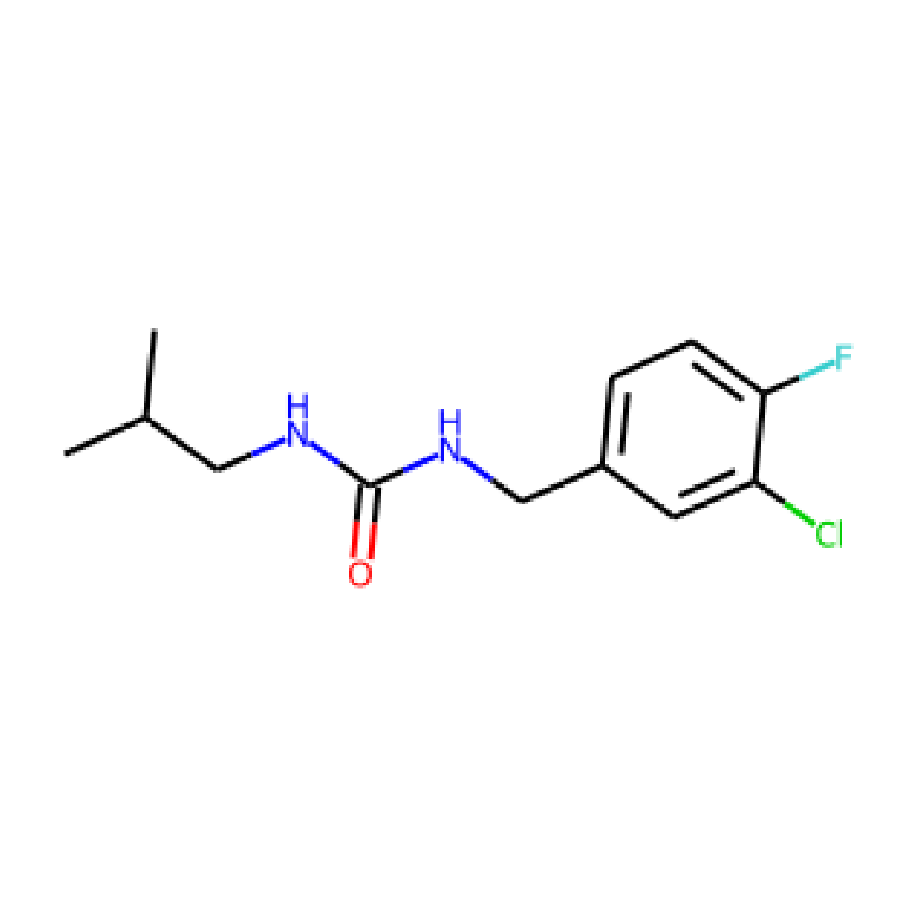}}& 
        {\includegraphics[width=0.185\textwidth]{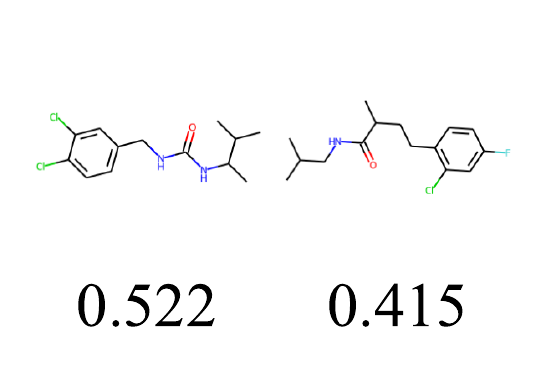}}&
        {\includegraphics[width=0.185\textwidth]{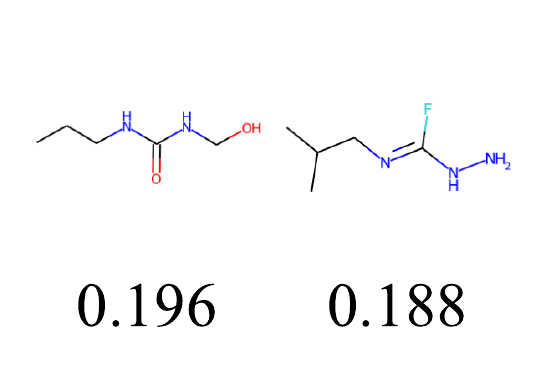}}&
        {\includegraphics[width=0.185\textwidth]{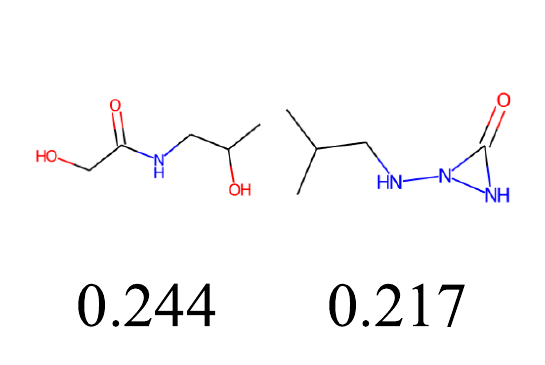}}&
        {\includegraphics[width=0.185\textwidth]{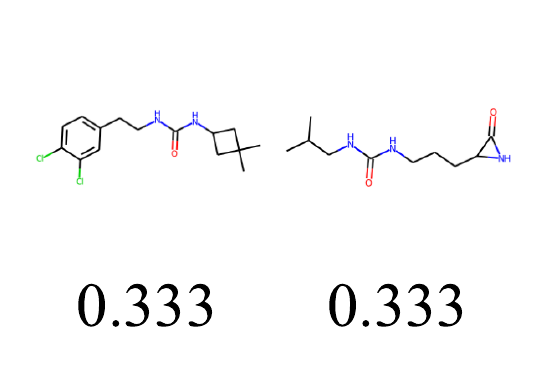}}\\
        {\includegraphics[width=0.10\textwidth]{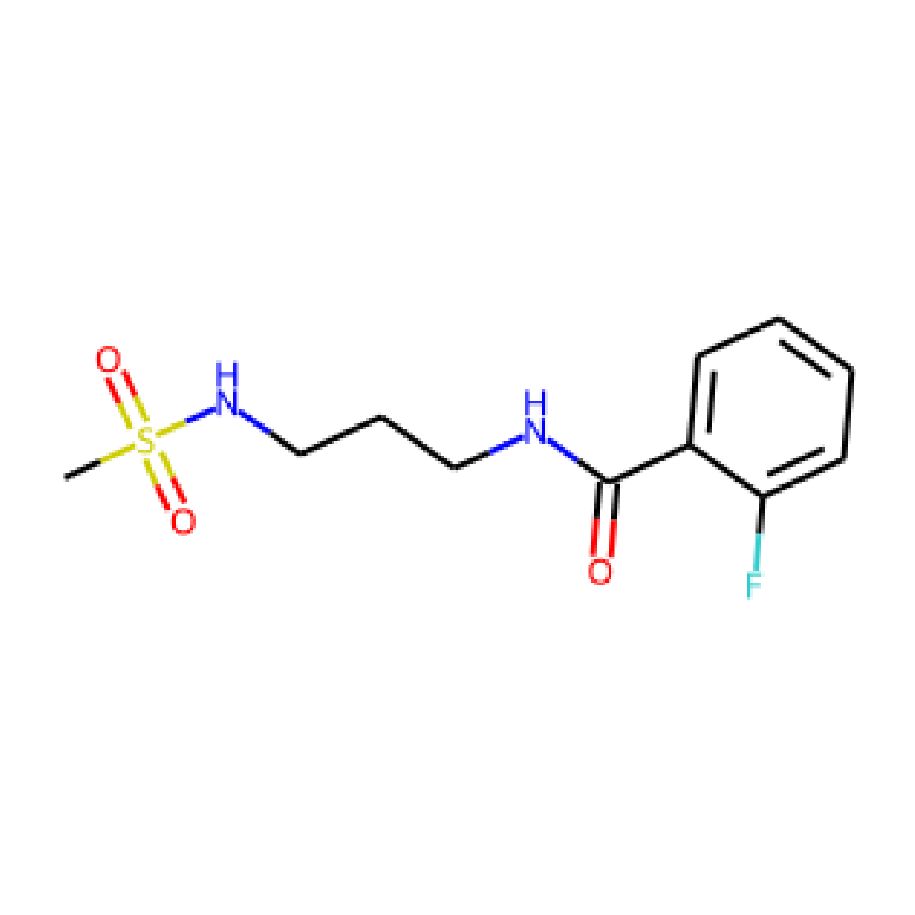}}& 
        {\includegraphics[width=0.185\textwidth]{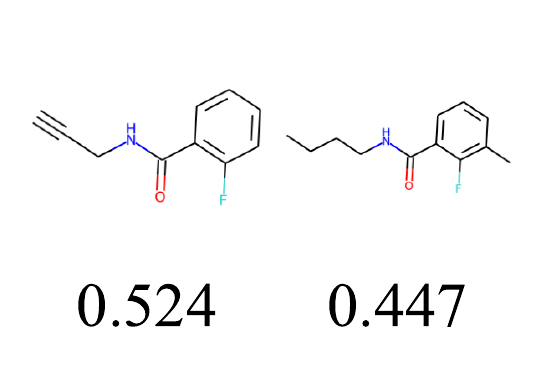}}&
        {\includegraphics[width=0.185\textwidth]{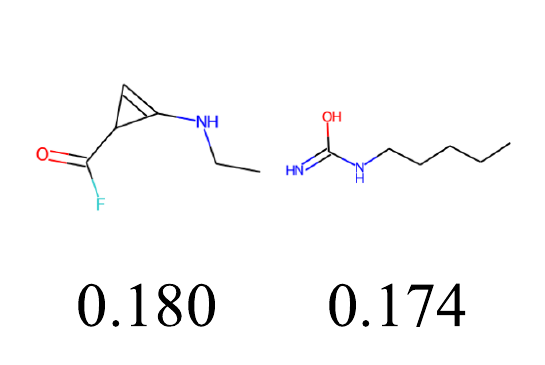}}&
        {\includegraphics[width=0.185\textwidth]{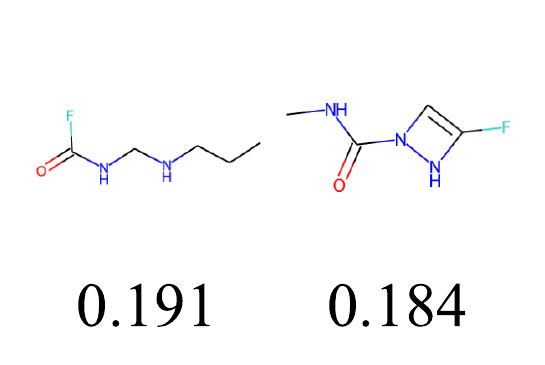}}&
        {\includegraphics[width=0.185\textwidth]{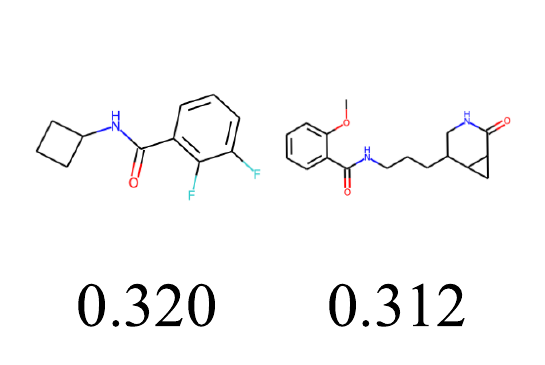}}\\
        {\includegraphics[width=0.10\textwidth]{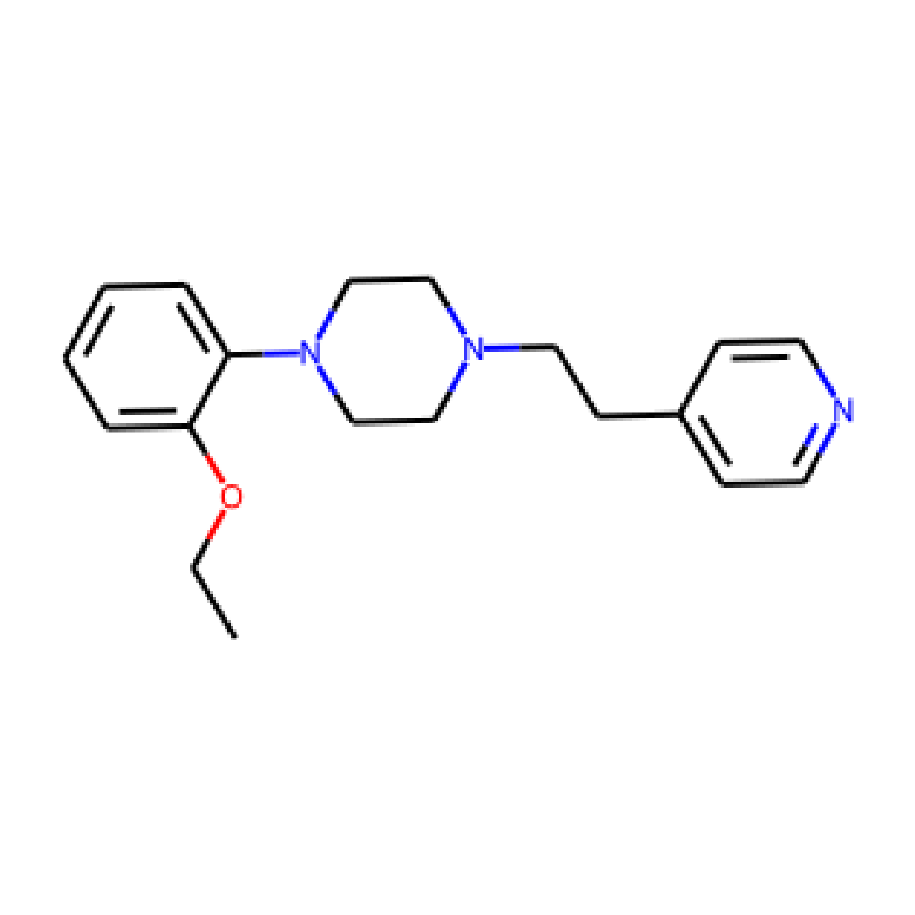}}& 
        {\includegraphics[width=0.185\textwidth]{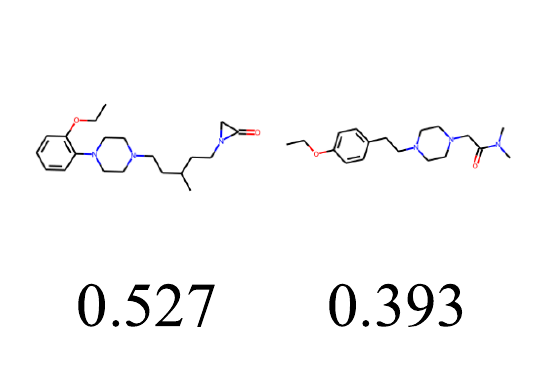}}&
        {\includegraphics[width=0.185\textwidth]{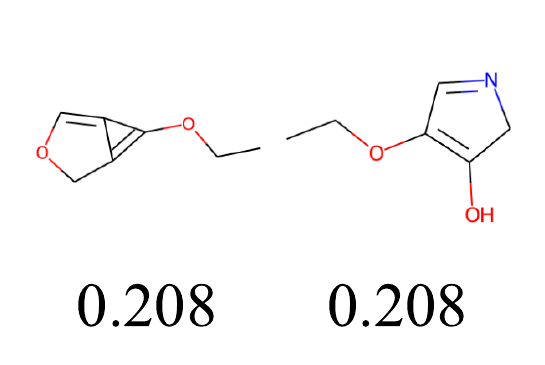}}&
        {\includegraphics[width=0.185\textwidth]{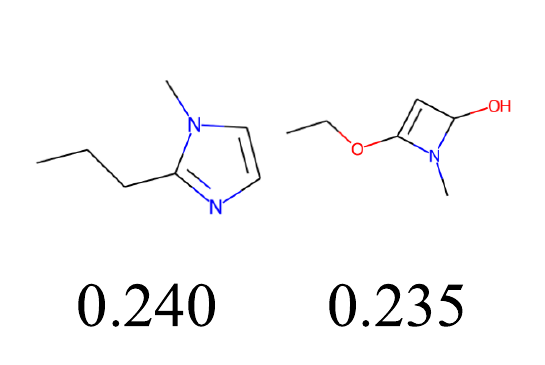}}&
        {\includegraphics[width=0.185\textwidth]{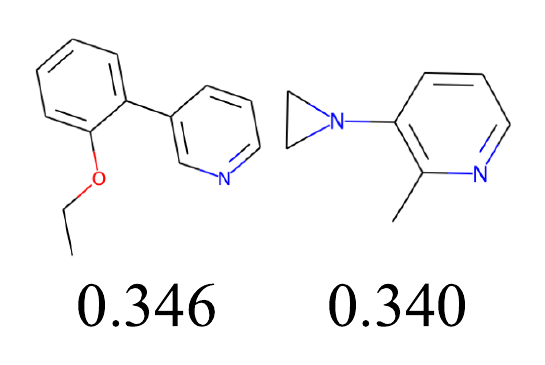}}\\
        {\includegraphics[width=0.10\textwidth]{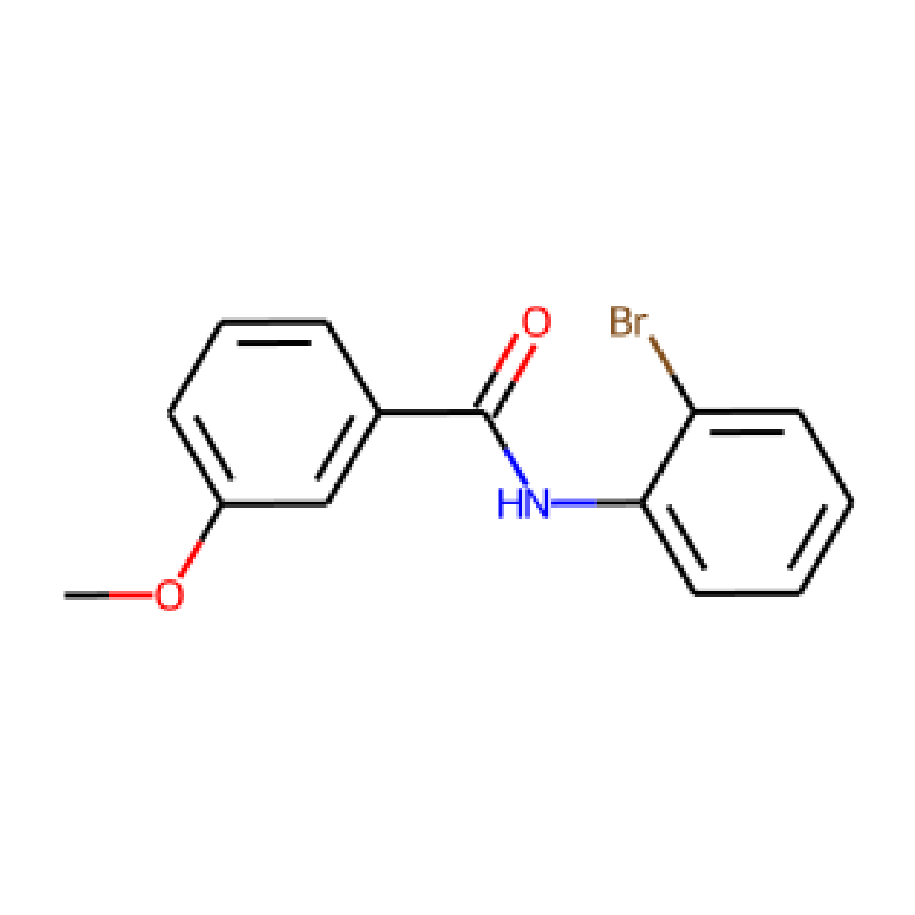}}& 
        {\includegraphics[width=0.185\textwidth]{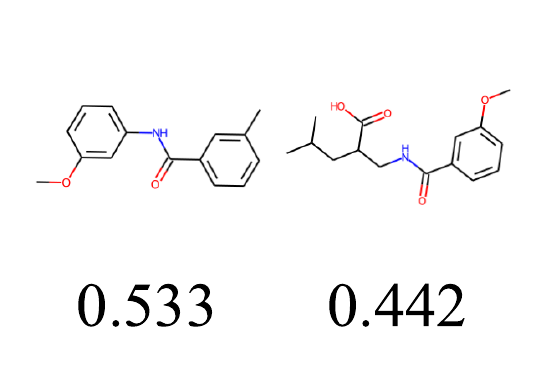}}&
        {\includegraphics[width=0.185\textwidth]{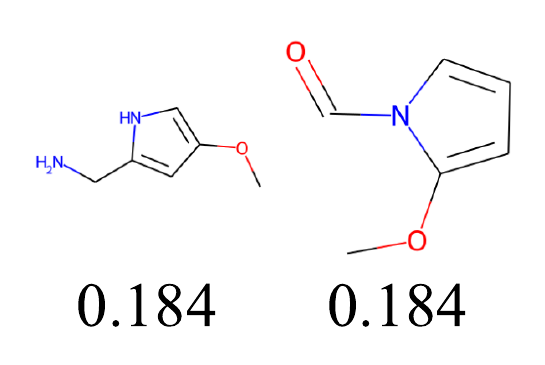}}&
        {\includegraphics[width=0.185\textwidth]{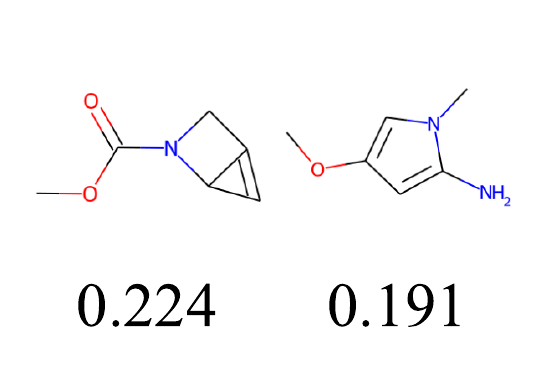}}&
        {\includegraphics[width=0.185\textwidth]{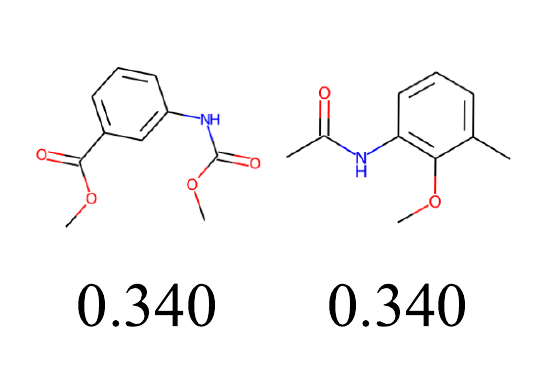}}\\
        {\includegraphics[width=0.10\textwidth]{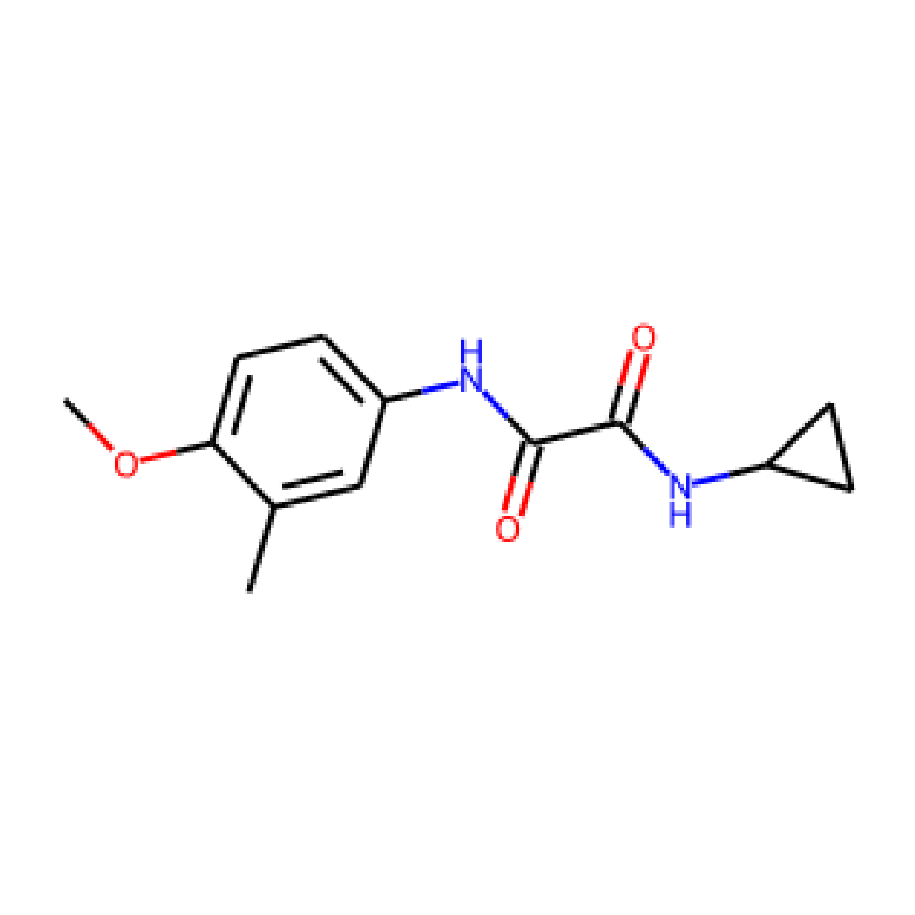}}& 
        {\includegraphics[width=0.185\textwidth]{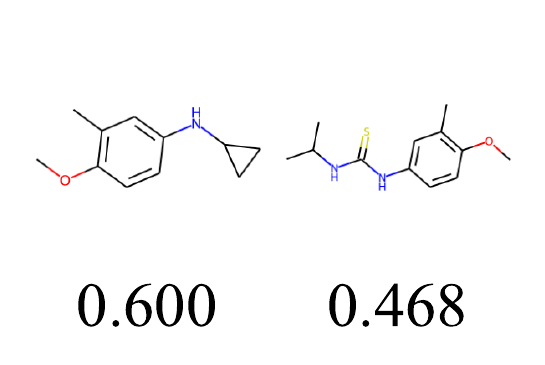}}&
        {\includegraphics[width=0.185\textwidth]{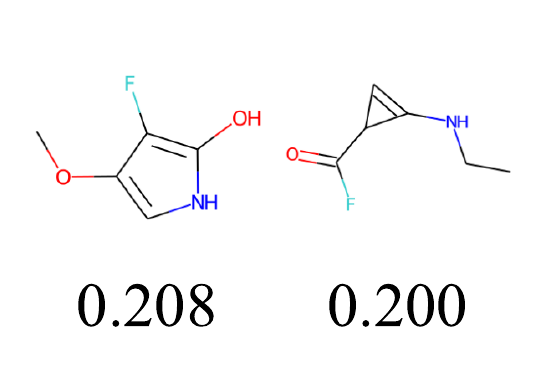}}&
        {\includegraphics[width=0.185\textwidth]{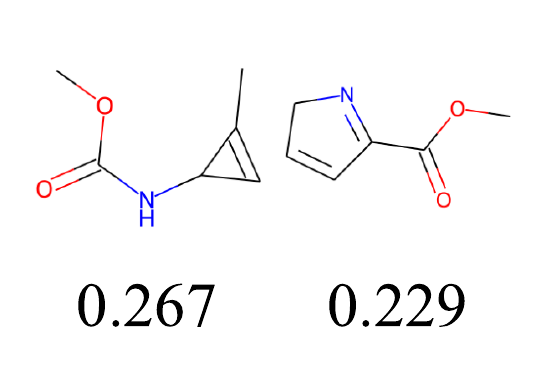}}&
        {\includegraphics[width=0.185\textwidth]{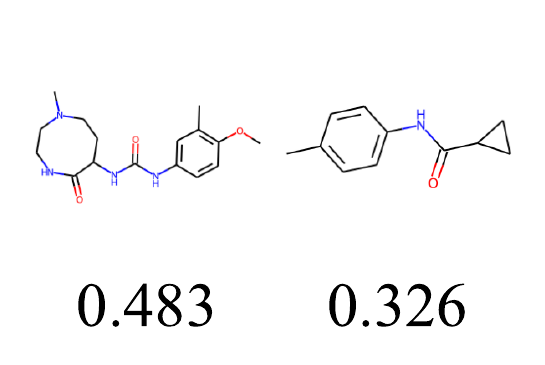}}\\
        {\includegraphics[width=0.10\textwidth]{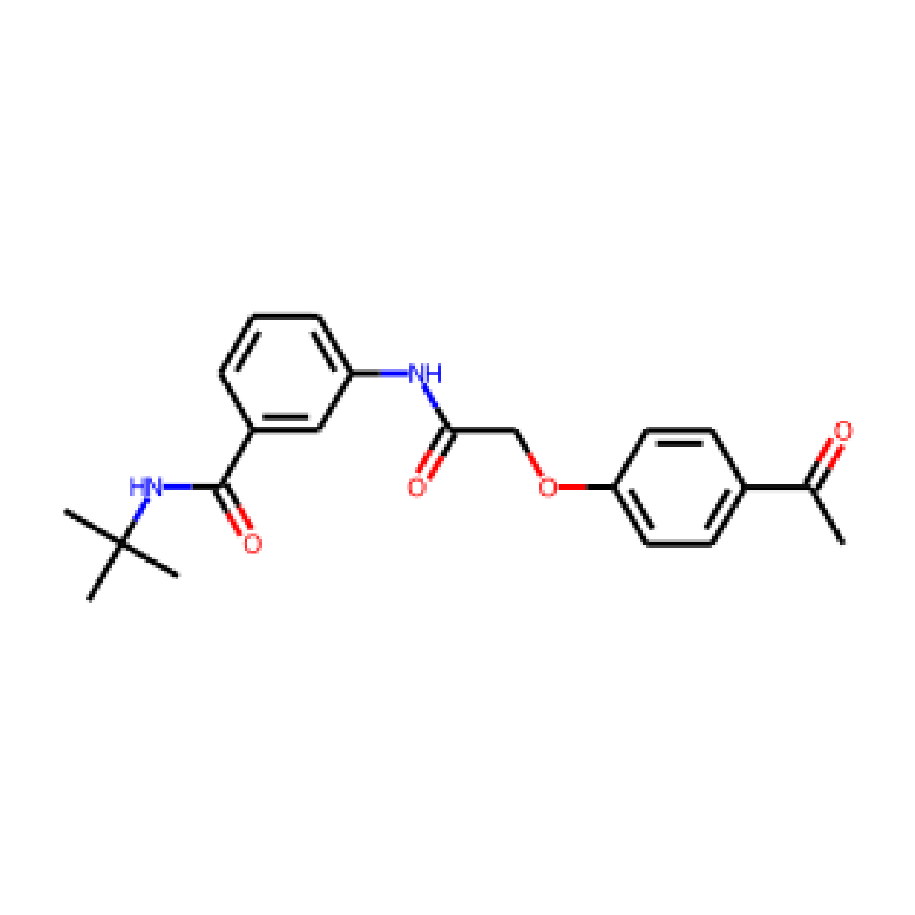}}& 
        {\includegraphics[width=0.185\textwidth]{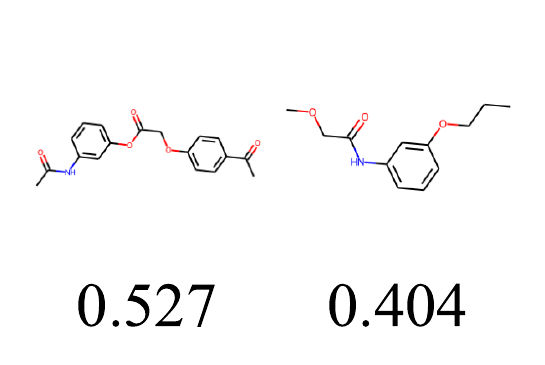}}&
        {\includegraphics[width=0.185\textwidth]{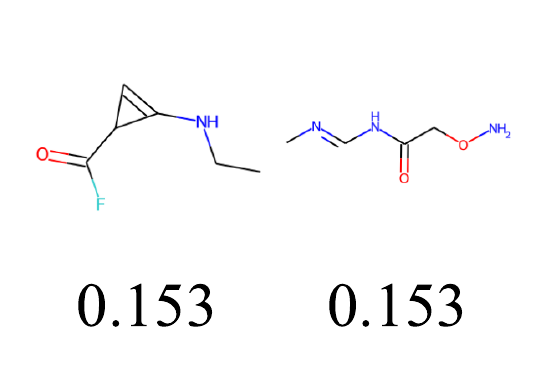}}&
        {\includegraphics[width=0.185\textwidth]{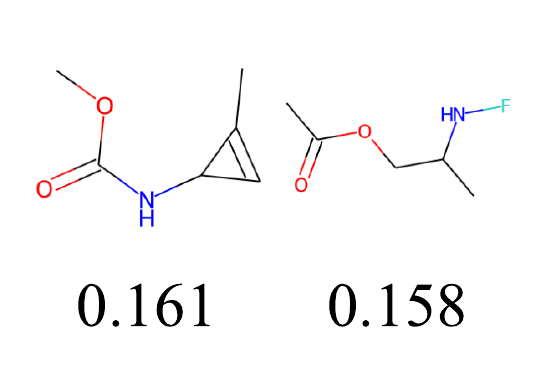}}&
        {\includegraphics[width=0.185\textwidth]{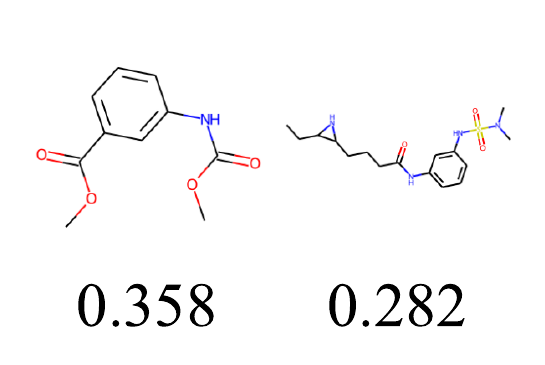}}\\
        {\includegraphics[width=0.10\textwidth]{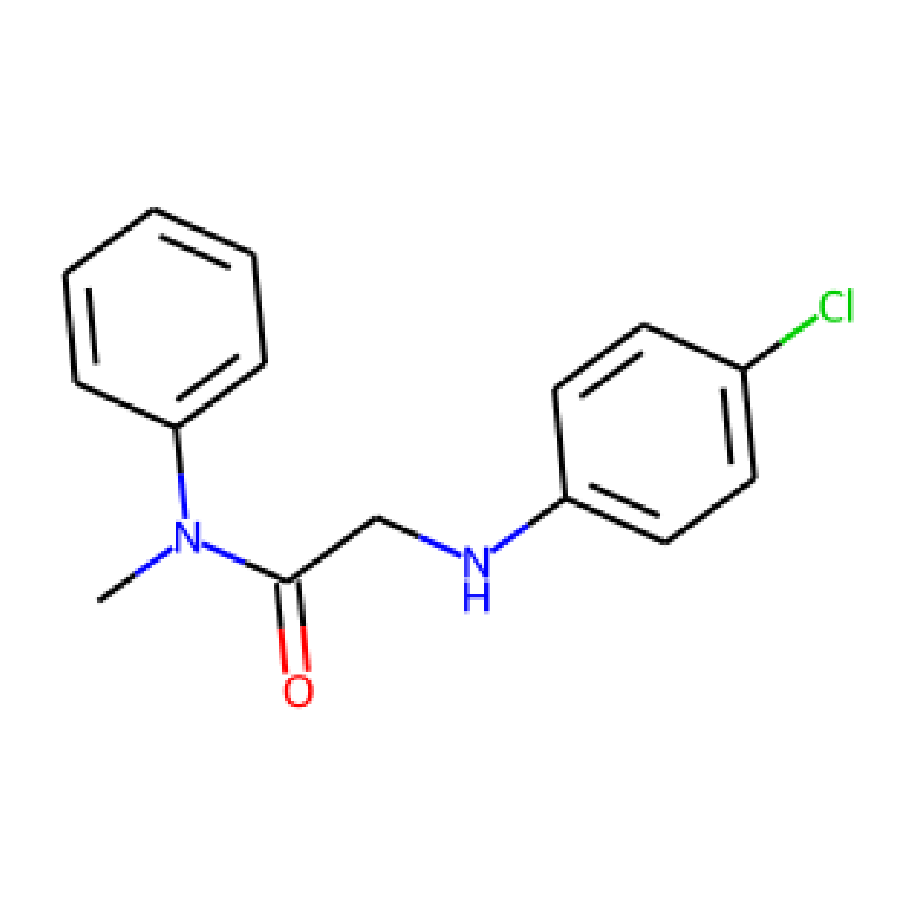}}& 
        {\includegraphics[width=0.185\textwidth]{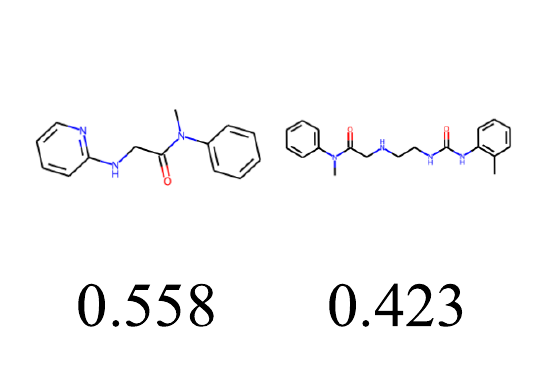}}&
        {\includegraphics[width=0.185\textwidth]{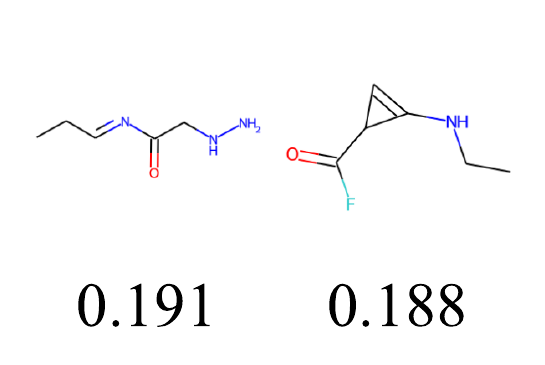}}&
        {\includegraphics[width=0.185\textwidth]{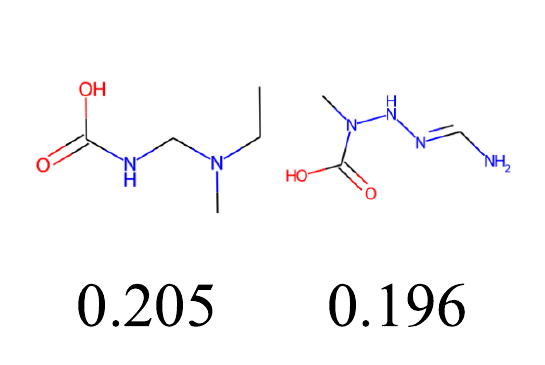}}&
        {\includegraphics[width=0.185\textwidth]{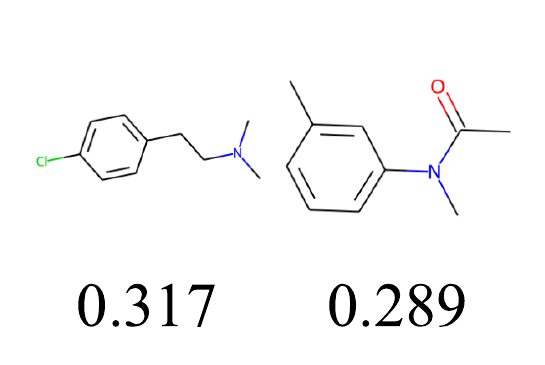}}\\
        
      \end{tabular}
      
    \caption{Extended result of molecule visualization on ZINC250K.}
    \label{fig:mol_vis_zinc}
\end{figure*}

Here we visualize the result for eight reference molecules randomly picked from the dataset. We visualize the top 2 generated samples ranked by the similarity score. Figure \ref{fig:mol_vis_qm9} and Figure \ref{fig:mol_vis_zinc} show the visualization results.

\end{document}